\documentclass[10pt]{article} 
\usepackage[preprint]{tmlr}

\usepackage{amsmath,amsfonts,bm}

\def\eqref#1{equation~\ref{#1}}

\def\1{\bm{1}}

\DeclareMathAlphabet{\mathsfit}{\encodingdefault}{\sfdefault}{m}{sl}
\SetMathAlphabet{\mathsfit}{bold}{\encodingdefault}{\sfdefault}{bx}{n}

\usepackage{hyperref}
\usepackage{url}
\usepackage{graphicx}
\usepackage{booktabs}
\usepackage{amsmath}

\makeatletter
\newcommand{\unnumberedfootnote}[1]{%
  \begingroup
    \let\@makefnmark\@empty
    \long\def\@makefntext##1{\parindent 1em\noindent ##1}%
    \footnotetext{#1}%
  \endgroup
}
\makeatother

\title{Replication Without Persistence in Hosted LLMs:\\
Measurement Sensitivity in Action-Time Belief Evaluation}

\author{\name Bhushan Kashinath Joshi \\[0.3em]
      \addr bjoshi2779@yahoo.com}

\def\month{MM}  
\def\year{YYYY} 
\def\openreview{\url{https://openreview.net/forum?id=XXXX}} 

\begin{document}

\maketitle

\begin{abstract}
Behavioural evaluations of hosted language models can vary because the
evaluated service, the measurement instrument, or both differ across runs. We
separate three validation questions: whether a prior finding recurs on fresh
data under its historical configuration (replication), whether the endpoint
changes when the evaluation-and-inference configuration is rebuilt under the
same identifier (measurement sensitivity), and whether the finding persists
across subsequently tested identifiers under one common instrument
(persistence). We study these questions in Regent Chess, a sequential
environment in which a hidden, mutable state is recorded exactly, allowing
stated beliefs to be scored against ground truth at action time; positive
endpoint values mean worse performance than a matched-uniform comparator. The
previously reported Gemini 3.1 Flash-Lite deficit recurs on fresh games under
its historical configuration ($+0.0530$, 95\% CI $[+0.0329,+0.0714]$). In a
back-to-back same-day H/R comparison under the same public identifier, the
model-minus-uniform endpoint is $0.0429$ lower under the rebuilt configuration
(95\% CI for the H-minus-R contrast $[+0.0182,+0.0667]$); all six
configuration components vary jointly, so no component is isolated. Under
rebuilt R, the prospectively frozen, interleaved same-window 4K comparison
reverses sign between Gemini 3.1 and Gemini 3.7, identifiers that differ in
release and product tier; additional descriptive and exploratory cells show
the same directional pattern. Any additional serving-period contribution
remains unresolved ($-0.0166$, $[-0.0483,+0.0157]$). Replication, measurement
sensitivity, and persistence can therefore yield different conclusions within
one evaluation, motivating explicit indexing of hosted-model behavioural
claims by tested identifier, serving period, measurement instrument, and
inference configuration.
\end{abstract}

\unnumberedfootnote{Generative AI tools were used in this work as assistive tools:
methodological critique and design discussion; implementation of
analysis, figure-generation and independent cross-checking code,
verified against frozen result artifacts; verification of literature
citations against primary sources; figure preparation; and manuscript
drafting and editing. They were not used to generate data. Scientific
decisions---including hypotheses, experimental freezes, endpoint and
classification-rule definitions, execution authorisation, and
interpretation of results---remained under human control;
responsibility for the final content likewise remains human.}

%

\section{Introduction}
\label{sec:intro}

A behavioural evaluation of a hosted language model is usually reported as
a property of a model. The finding is
attached to a public identifier, and the identifier is treated as naming
the thing that was measured. But an identifier addresses a service, not an
artefact. The same name can be served from a changed backend, and the same
name can be measured through a different harness, prompt assembly,
transport, or inference configuration. Neither of those is a defect of the
original study; both mean that a behavioural claim indexed only to a name
is under-specified, and that anyone rerunning it is not necessarily
measuring the same object.

This matters most for capabilities that are themselves fragile. Work on
confidence and calibration has established that language models can be
poorly calibrated, systematically overconfident, and highly variable in
calibration quality across families and versions
\citep{kadavath2022know,wang2026calibrating,ffrenchconstant2026confidencebench,
leng2025taming,kim2026same}. Almost all of that evidence comes from static
settings, where a model answers a question or labels an instance and
reports confidence in that single act, so the scored answer and the stated
confidence share one channel. Considerably less is known about belief
quality in interactive settings, where evidence accumulates over time and
belief is elicited at the specific moment an action is taken --- the moment
at which a stated belief, if it is wrong, actually costs something.

Rerunning such a finding raises three questions that are routinely
collapsed into one. \emph{Replication} asks whether the finding recurs on
newly generated data when the original measurement configuration is rerun.
\emph{Measurement sensitivity} asks how much the endpoint moves when the
configuration is rebuilt while the model identifier and serving period are
held fixed. \emph{Persistence} asks whether the finding continues to
characterise subsequently tested identifiers evaluated through one common
instrument. These are different questions with different comparisons, and
a single number cannot answer more than one of them. The central empirical
content of this paper is that they can come apart: the same finding
replicates under its own historical configuration and fails to persist
across the next identifiers tested under a rebuilt one.

We examine all three in Regent Chess \citep{joshi2026confident}, a chess
variant in which royal status can be transferred secretly between a
player's own pieces during play. The piece holding royal status --- the
\emph{Regent} --- is the piece an opponent must capture to win, so its
identity is a latent fact that is consequential, that changes at moments
its owner chooses, and that is already determined by prior events at every
scored decision rather than resolved afterwards. The environment's engine
records the true Regent identity exactly and independently of anything
either player reports, so a model's stated probability distribution over
candidate squares can be scored against ground truth at the moment it
acts, while it is playing to win. Evidence about the hidden state
accumulates through ordinary play, which makes the task tracking rather
than static recall. We use the environment as a measurement instrument,
not as a test of chess strength.

The study starts from a published behavioural claim: in that earlier work,
one hosted model's action-time belief quality was severely degraded, with
only 1 of 62 high-confidence capture beliefs correct across two runs
\citep{joshi2026confident}. We first rerun that study's own measurement
configuration on newly generated games in a later serving period, which is
a fresh-data replication. We then hold the public model identifier and the
serving day fixed and substitute a rebuilt measurement configuration, which
isolates the measurement question from the temporal one as far as the
design allows. Finally we test persistence across subsequently tested
identifiers under that one rebuilt configuration, in a designed $2\times 2$
crossing two identifiers with two generated-token ceilings, three of whose
cells were sealed before their outcomes were known while the fourth had
already been observed and is reported descriptively, plus an exploratory
extension to a third identifier released during execution and a
cross-family transport.

The historical deficit replicates directionally on fresh games under its
own configuration ($+0.0530$, 95\% CI $[+0.0329, +0.0714]$, relative to a
matched-uniform comparator; positive means worse than the comparator).
Within one day and under the same identifier, substituting the rebuilt
configuration reduces that deficit by $0.0429$ ($[+0.0182, +0.0667]$).
Under the rebuilt configuration, the originally tested identifier still
shows the deficit at both token ceilings ($+0.0242$ and $+0.0262$), while
the next identifier tested reverses the endpoint direction at both
($-0.0131$ and $-0.0104$), as does the further exploratory release. Any
additional serving-period contribution remains unresolved at the available
precision ($-0.0166$, $[-0.0483, +0.0157]$).

We are deliberate about what these results are not. They are not a causal
claim about model versions: the tested identifiers differ in release and in
product tier at once, and the design randomises neither, so the contrast is
between tested identifiers rather than an isolated version effect. They are
not a demonstration of backend drift: the historical-versus-reconstructed
difference remains jointly indexed by configuration and serving period, and
the same-day control associates the endpoint shift with a six-component
configuration bundle without identifying any single component within it.
They are not a claim that the task is solved in the later conditions, where
absolute error remains substantial. And they are not a general claim about
benchmarks: all of this comes from one sequential hidden-state environment,
one fixed opponent, and one provider lineage plus a single cross-family
transport.

\paragraph{Contributions.} (i) A fresh-data replication of a previously
published action-time belief finding, rerun under that study's own
measurement configuration in a later serving period. (ii) A
measurement-sensitivity control that holds model identifier, output ceiling,
opponent, game horizon, seed schedule, and scorer fixed within a single day
and varies only the configuration bundle. (iii) An across-identifier
persistence test under one common rebuilt instrument, as a designed
$2\times 2$ in which cells A--C were sealed before their outcomes were
known while the earlier-observed D cell is reported descriptively, with a
separately reported exploratory extension and a cross-family
transport. (iv) An
elicitation-method analysis: three increasingly constrained full-support
protocols, each prospectively frozen with its own decision rule, which
establish why the retained top-$k$-plus-residual instrument was adopted and
what it costs. (v) A behavioural decomposition conditioned on states where
the hidden target was legally actionable, with depth and state-richness
robustness checks. (vi) Independent reimplementations sharing no scoring
code, built from the frozen endpoint definitions, which reproduced every
reported quantity; all quantities reported here derive from a frozen,
independently audited result set.

\section{Action-Time Hidden-State Measurement}
\label{sec:measurement}

Scoring a stated belief at the moment of action requires four properties of
the environment, and they are jointly restrictive. Evidence about the
hidden fact must accumulate sequentially, so that the task is tracking.
The fact must already be determined by prior events at each scored
decision, rather than being resolved after the model commits, so that a
probability statement about it has a truth value when it is made. The
environment must record that fact exactly and independently of anything the
model reports, so that scoring does not depend on a modelling assumption
about ground truth --- which is not guaranteed in interactive hidden-state
settings more broadly. And the fact must be able to change during play, so
that a correct answer cannot be recalled from the initial configuration.
Regent Chess satisfies all four.

Only a little of the game matters here. Each player begins with royal
status on their own king and may, during play, transfer it secretly to any
of their own live pieces; this is a \emph{Crown Shift}. The opponent is
never told that a shift occurred or where royal status now sits, so the
identity of the opponent's Regent is hidden and mutable. Capturing the
opponent's Regent wins immediately, which is what makes the latent state
consequential rather than incidental: a shift onto an attacked piece can
lose the game at once, so shift timing is itself a decision under
uncertainty. A model observes the full board and move history, as in
standard chess, together with its own royal status and mode; it never
receives the opponent's royal status, which it must infer from observable
play. The engine tracks the true Regent identity throughout, independently
of what either player is told, so it is exactly recoverable afterwards for
scoring. Full rules are in Appendix~\ref{app:rules}.

On every real turn the harness issues two separate calls. One selects the
move to play. The other elicits a probability distribution over which of
the opponent's live pieces currently holds royal status, conditioned on the
same game state but \emph{not} on the move just chosen. Keeping them apart
is what makes it possible to ask whether a model's stated belief and its
selected action agree at all --- a question that is structurally
unanswerable wherever the scored answer and the stated confidence are the
same output. Belief is elicited as an explicit list of the top-$k$ named
candidate squares with their probabilities, plus a single literal residual
mass for every unnamed live piece (``OTHER''), rather than as an exhaustive
distribution over all live pieces. Section~\ref{sec:elicitation} reports
the protocol studies behind that choice. One consequence is load-bearing
throughout: the model chooses its own named candidate set, and therefore
chooses how many scored events each ply contributes.

Belief quality is scored two ways. R1, the primary endpoint, is a binary
Brier-loss difference over named-candidate events. For each eligible
well-formed ply at board ply $\geq 8$ and each explicitly named candidate
square $s$ carrying stated probability $p$, the model's loss is
$R1_{\mathrm{model}} = \operatorname{mean}_{(s,p)}(p - y_s)^2$, where
$y_s = 1$ if and only if $s$ is the true Regent square. The matched
comparator is evaluated on the identical events with
$p_{\mathrm{uniform}} = 1/N$, where $N$ is the number of live opponent
pieces at that ply, and the reported quantity is
$\Delta_{R1} = R1_{\mathrm{model}} - R1_{\mathrm{uniform}}$. Stated
probabilities enter exactly as elicited: there is no renormalisation, no
clamping, and no repair of a distribution that fails validity. A response
is scored only if it parses to a valid candidate distribution whose total
stated mass lies within $0.02$ of $1$; responses that do not are recorded
as non-well-formed and excluded from scoring rather than mended
(Appendix~\ref{app:wellformed}). Three properties of R1 follow from its
construction. Because the model selects the named set, it also selects the
event count per ply. Because $y_s = 0$ for every named square whenever the
true Regent is not among them, a ply on which the true state falls in OTHER
contributes only zero-labelled events, so R1 does not directly penalise
omitting the true state. And because the event set is model-selected, the
aggregate is support-dependent.

R2 is complementary and addresses the omission that R1 does not. It is a
Brier sum over the full elicited partition, comparator $1/N$ on each named
square and $(N-k)/N$ on OTHER, so placing the true state outside the named
candidates is penalised directly. R2 magnitudes are support-dependent in
the same way as R1's, and R2 is reported alongside R1 rather than in place
of it. Both endpoints are scored on every eligible belief ply past the
canonical cutoff and neither requires that an actionable capture of the
hidden target exist at that ply; conditioning on actionability is used only
for the secondary analysis in Section~\ref{sec:decomposition}, which
defines its own population.

The interpretive limitation this imposes should be stated exactly, because
it bounds how the endpoint may be read across conditions. The binary Brier
rule is proper conditional on a fixed event set. The aggregate endpoint,
however, is computed over an event set the model chose, and mean
named-candidate support differs materially between the tested identifiers.
Cross-condition R1 \emph{magnitudes} at materially different support are
therefore not directly comparable as global belief-quality magnitudes. This
does not affect the prespecified status of the sign and classification
endpoint, which is what the confirmatory design rests on, but every
magnitude comparison in this paper remains conditional on the
model-selected event set.

From $\Delta_{R1}$'s 95\% interval against zero, a frozen classification
rule assigns one of three labels: \textsc{replicates} when the interval
lies entirely above zero, meaning the model scores worse than the matched
comparator and so agrees in direction with the historical degradation;
\textsc{contradicts} when it lies entirely below, meaning the model scores
better and the direction is reversed; and \textsc{unresolved} when it spans
zero. One point about the first label is easy to misread and is therefore
stated explicitly here: \textsc{replicates} is an \emph{endpoint-status
label} denoting directional agreement of an interval with a sign. It is not
a study-design claim that a replication occurred. We retain the label
unchanged because it was preregistered, and we do not reuse it to describe
study design.

We follow the standard distinction between two senses of repeating a study
\citep{nasem2019reproducibility}. \emph{Computational reproducibility} is
recovering reported results from archived data and specifications ---
including by an independent reimplementation --- and is a property of the
analysis pipeline. \emph{Replication} is a fresh study on newly generated
stochastic data. In this paper the historical configuration rerun in a
later serving period is a fresh-data replication, because new games were
generated; the factorial and extension conditions are evaluations under the
rebuilt configuration and are never described as replications of the
historical experiment. Neither sense requires that a hosted model produce
identical generations, which it cannot be made to do.

\section{Choosing the Belief-Elicitation Instrument}
\label{sec:elicitation}

The elicitation format is a scientific choice rather than an
implementation detail, because it determines what the endpoint is computed
over. The instrument retained here --- top-$k$ named candidates plus a
single OTHER residual --- is not the obvious choice. An exhaustive
distribution over every live opponent piece would be cleaner: the support
would be fixed by the position rather than by the model, so the
support-dependence limitation of Section~\ref{sec:measurement} would not
arise, and omission could not occur. We report here why that cleaner
instrument was not adopted. The reason is empirical, not aesthetic, and it
was established before the factorial grid was designed, in three
increasingly constrained protocol studies.

The three protocols share a design that makes them comparable. Belief
elicitation is stateless, so both formats in a given comparison could be
issued at the identical pre-move state within one game, making each
comparison paired rather than a contrast between two samples. Each
protocol was prospectively frozen with its own read order, decision rule,
and list of quantities its ruling was forbidden to inspect, and each ran
against the same fixed heuristic opponent with the rest of the evaluation
design unchanged. All three were run against a single tested configuration:
the public identifier \texttt{gemini-3.1-flash-lite}, elicited through the
pre-factorial pilot harness over an OpenAI-compatible endpoint with the
disambiguated own-state presentation at a 4{,}096-token generated-output
ceiling. Their evidence is scoped to that identifier under that harness,
and not to the rebuilt configuration R under which the grid was later run.

\paragraph{Protocol B: self-enumerated full support.} The model is asked to
reconstruct the live candidate set itself and assign a probability to every
member, with no residual category. Against a non-inferiority margin of
$-0.15$ fixed before execution, the prespecified ply 1--25 band (334
checkpoints across 24 games) showed well-formedness falling from 90.4\%
(302/334) under top-$k$ to 24.6\% (82/334) under full support, a paired
difference of $-0.6587$ with 95\% CI $[-0.7157, -0.6032]$. The entire
interval lies below the margin, which is a formal \textbf{NO-GO}. The two
later bands (ply 26--50 and ply 51+) were \textsc{unresolved} at small
sample sizes, so the decision rests on the prespecified 1--25 band alone,
not on a pooled result. The failure mode was specific rather than generic:
nearly every non-well-formed full-support response named a candidate that
was not live or omitted one that was, rather than failing to parse at all
(full taxonomy, Appendix~\ref{app:elicitation}). What collapsed was the
model's ability to enumerate the live candidate set, not its ability to
produce structured output.

\paragraph{Protocol C: roster-supplied positional elicitation.} If
enumeration is the difficulty, supplying the roster should remove it. The
harness therefore provided the exact live-candidate roster and the model
returned a probability vector aligned to it. Structural feasibility
improved substantially. But a validation that permuted only the order in
which the roster was presented --- holding board state, history, and
candidate set fixed --- found that on every one of the three genuinely
informative states, variation across orderings exceeded variation between
replicates at a fixed ordering, by factors of $2.00\times$ to
$15.00\times$. The roster-supply insight was retained; the positional
response encoding was not, because a response that depends on the order of
an irrelevant presentation is not a measurement of belief. That motivated
binding each probability to a named candidate rather than to a position.

\paragraph{Protocol D: identity-anchored roster supply.} Each probability
is bound explicitly to a named candidate, which removes transcription
between position and identity as a failure mode. This was assessed on two
distinct instrument properties, and the distinction matters for how the
results read. D-F1 tested structural feasibility against concurrent
top-$k$: the paired difference was $+0.2235$, 95\% CI
$[+0.0824, +0.3647]$, entirely above the $-0.15$ margin, a formal
\textbf{GO}. D-F2 then estimated presentation stability: the mean
total-variation distance between responses to states differing only in the
order in which hypotheses were presented was $0.1809$, 95\% CI
$[0.1247, 0.2395]$.

\begin{figure}[t]
\centering
\includegraphics[width=0.99\linewidth]{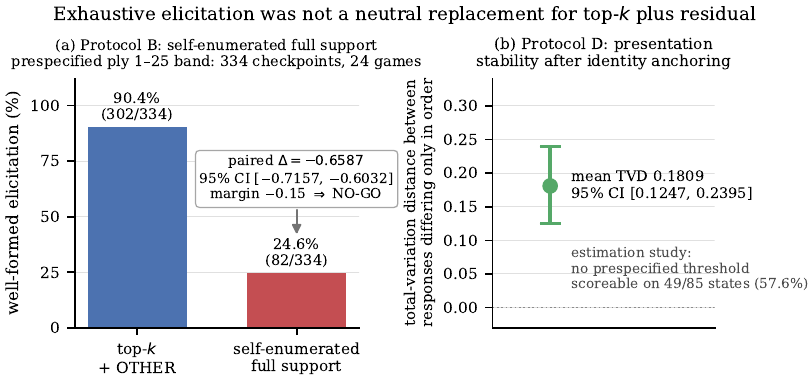}
\caption{\textbf{Why exhaustive elicitation was not a neutral replacement
for top-$k$ plus residual.} \textbf{(a)} Protocol B, in the prespecified
ply 1--25 band: the well-formed elicitation rate under top-$k$ plus
residual against the rate under self-enumerated full support, with the
paired difference, its game-clustered interval, and the non-inferiority
margin fixed before execution. The entire interval lies below that margin,
which is a formal \textbf{NO-GO}; the decision rests on this prespecified
band alone, the two later bands having been \textsc{unresolved} at small
sample sizes. \textbf{(b)} Protocol D-F2: the mean total-variation distance
between responses to states differing only in the order in which
hypotheses were presented, with its 95\% interval. \textbf{D-F2 was an
estimation study with no prespecified pass or fail threshold, and is not a
failed test;} it met the precision target it declared in advance, and the
interval's position is evidence about the size of a presentation effect
rather than a verdict against a criterion. The estimate is conditional on
the 49 of 85 states that yielded four independently well-formed responses;
it is \emph{not} a lower bound, and no direction of that selection effect
is identified. It also does not reverse D-F1's \textbf{GO} on structural
feasibility ($+0.2235$, 95\% CI $[+0.0824,+0.3647]$ against the same
margin), which tests a different instrument property. Protocol C is
described in the text and is deliberately not shown.}
\label{fig:elicitation}
\end{figure}

D-F2 was an estimation study rather than a test against a threshold, and
it does not reverse D-F1: the two measure different instrument properties,
so a single call can be reliably well-formed while repeated exhaustive
elicitation at the same state remains brittle. Figure~\ref{fig:elicitation}
states the qualifications that bound the estimate. What it supports is a
practical measurement judgement: the residual presentation sensitivity was
large enough that we did not treat identity-anchored full support as
sufficiently stable to carry downstream belief-quality analysis under the
tested configuration. Development stopped there, under the scope rule
frozen in advance; no fourth protocol was designed.

Read together, the three protocols establish a narrow but decisive claim:
under this model and this evaluation design, exhaustive elicitation was not
a neutral drop-in replacement for top-$k$-plus-residual. Two distinct
obstacles appeared --- self-enumeration of the candidate set failed
outright, and once the set was supplied, the response was sensitive to
presentation --- and only the second was substantially mitigated by
redesign.

The retained instrument is therefore a practical measurement choice, not an
ideal one. It carries the support dependence and residual-mass costs that
Section~\ref{sec:measurement} sets out, and one further cost of its own:
because top-$k$ truncates, the instrument cannot by itself distinguish a
candidate the model ranked below the cut from one it never entertained.
What it does deliver is a
format that this class of model can satisfy at a high rate in real games,
issued identically across every condition, with validity checked rather
than repaired. The scope of the protocol evidence is correspondingly
narrow: it characterises one model and one configuration, a different
model might enumerate reliably, and each protocol's frozen ruling was
forbidden to inspect Regent truth, correctness, proper scores, or acted-on
probability --- so these studies say nothing about what belief-quality
result an exhaustive protocol would have produced. Full protocol detail,
sample counts, and failure taxonomies are in
Appendix~\ref{app:elicitation}.

\section{Experimental Design and Provenance}
\label{sec:design}

Two measurement configurations appear in this paper and the distinction
between them organises everything that follows. The \emph{historical
configuration} (H) is the harness of the earlier study, which predates this
work; it is the configuration under which the original finding was
produced, and rerunning it on newly generated games is what makes a
fresh-data replication possible. The \emph{rebuilt configuration} (R) is a
native-API instrument built for this study, and it is the common instrument
under which every factorial and extension condition was evaluated. Because
the historical and reconstructed evaluations differ in measurement
configuration \emph{and} in serving period at once, their difference alone
identifies neither source. That is the reason for the control described
next.

\paragraph{The same-window configuration control.} We reran H and R in the
same serving window under the same public model identifier, holding fixed
the identifier itself, the 4{,}096-token generated-output ceiling, the fixed
heuristic opponent, a 150-ply game horizon, the shared environment-seed
schedule, balanced colour assignment, and the common scorer, at 100 games
per arm. Only the configuration bundle varied. Two features of the design
bound its interpretation. First, the arms ran \emph{back-to-back on one
day} rather than randomised or interleaved, so residual intra-window
serving variation cannot be excluded. Second, H and R differ as a bundle of
six components, not as a manipulated factor: the transport (an
OpenAI-compatible endpoint versus the provider's native API), the
elicitation mode (a combined move-and-belief call versus two separate
calls), the presentation of the model's own state, the thinking setting
(none versus a fixed medium level), whether sampling parameters are
forwarded, and the revision of the rules document supplied in the prompt.
Under the reporting scheme used here, the H/R bundle crosses both the
\emph{measurement-instrument} and \emph{inference-configuration} coordinates
of Section~\ref{sec:discussion}: transport, elicitation mode, own-state
presentation and the rules-document revision belong to the instrument,
while the thinking setting and sampling-parameter forwarding belong to the
inference configuration. That grouping is an operational classification
applied after the fact, not a decomposition of the treatment; all six
components varied jointly. The consequence for inference is that the
endpoint difference is \emph{associated with the evaluation and inference
configuration bundle as a whole}: neither an individual component nor
either coordinate's sub-bundle is isolated by this design, and we do not
attribute the shift to any one of them.

\paragraph{What the rules-document component is not.} The last component in
that bundle could be mistaken for a change in the game, so we state what it
is: the two rules-document revisions differ only in a trailing version
block and contain no differing normative text, so the gameplay rules were
identical across the arms.

\paragraph{The core grid.} The persistence question is addressed by a
$2 \times 2$ factorial crossing two Gemini identifiers with two
generated-token ceilings, at 150 games per cell: cells A and B are
Gemini 3.1 Flash-Lite at the 4{,}096- and 16{,}384-token ceilings, and
cells C and D are Gemini 3.7 Flash at the same two ceilings. The opponent,
environment seeds, elicitation architecture, scoring rules, and stopping
criteria are fixed across all four cells. Two exploratory cells, E and F,
extend the design to Gemini 3.8 Flash --- an identifier released during
execution of the core grid --- at the same two ceilings.

The provenance of these cells is not uniform, and the differences change
what each one can support. Cell D was executed first: it ran in the
immediately preceding serving window and completed before the A/B/C
interleaved window opened, and it is what motivated building the grid. Its
outcome therefore preceded the freeze of the classification rule, so its
label is descriptive and is not prospectively confirmatory. Cells A, B, and C, together with the
classification rule itself, were frozen and sealed before any of their
scientific outcomes were known, and the three ran on interleaved schedules
within one serving window. The strongest prospective cross-identifier
comparison available is therefore \textbf{A versus C}: the same output
ceiling, the same rebuilt configuration, both prospectively frozen, and
interleaved execution. Comparisons involving D cross serving windows and
are reported as such. The Gemini 3.8 extension was initially frozen at 150
games per cell before any call to that identifier; after E's first 75-game
block completed, only prespecified operational information --- runtime,
cost, transport behaviour, and output-ceiling pressure --- was inspected,
and on that basis alone, before cell F began, the design was amended to 75
games per cell and downgraded to exploratory. No scientific endpoint,
belief content, or correctness signal from either cell was inspected at
that point. E therefore ran under the original confirmatory plan and F
entirely under the amended exploratory one, and neither carries
confirmatory weight anywhere in this paper.

One confound in the grid cannot be removed by design and is stated
wherever the comparison appears. Gemini 3.1 Flash-Lite and
Gemini 3.7/3.8 Flash differ in release \emph{and} in product tier, with
whatever difference in served capacity that entails. The design randomises
neither, nor the trajectory distribution each identifier induces through
its own play. Every cross-identifier statement in this paper is therefore a
statement about \emph{tested identifiers}, not about an isolated model
version.

\paragraph{The opponent.} All interactive conditions used the same fixed
heuristic opponent, referred to throughout as Tier-1, at its medium
setting: static one-ply move evaluation with stochastic selection among
its top three scored moves. The
label denotes a configuration, not an externally calibrated chess-strength
level, and it was fixed before any condition's results were known. Its
Crown-Shift policy is relevant to how predictable the hidden state is: it
shifts when its own Regent is attacked, and otherwise at a per-game
first-shift move drawn uniformly from moves 4--12 and thereafter with
15\% probability per turn, choosing the target by a deterministic safety
heuristic that favours unattacked, well-defended, non-central, and less
conspicuous pieces. Hidden-state timing is thus only partly
state-responsive, which bounds how far the latent state can be predicted
from public play at all --- a bound that applies to any predictor,
including the reference tracker discussed in
Section~\ref{sec:comparator}. Opportunity density for the actionability
analysis is likewise opponent-dependent, so results characterise this fixed
regime rather than an opponent-invariant property of a model.

\paragraph{Cross-family transport.} One condition sits outside the
factorial and outside its classification scheme. A prospectively frozen
cross-family transport tests whether the reversal is confined to the Gemini
lineage: GPT-5.6, accessed as the transport identifier ``Terra'' over a
different provider API at the 16{,}384-token ceiling, with the same fixed
opponent, the same elicitation architecture, and the same 75 environment
seeds as cell F. Its sample size was fixed prospectively rather than by an
outcome-dependent stopping rule. It is not simply another R cell: it
preserves R's elicitation architecture over a provider-specific
implementation with its own request shaping and reasoning configuration, so
it is transport-analogous to R rather than transport-identical, and we
report it separately and never under the factorial's classification labels.

\begin{figure}[p]
\centering
\includegraphics[width=\linewidth]{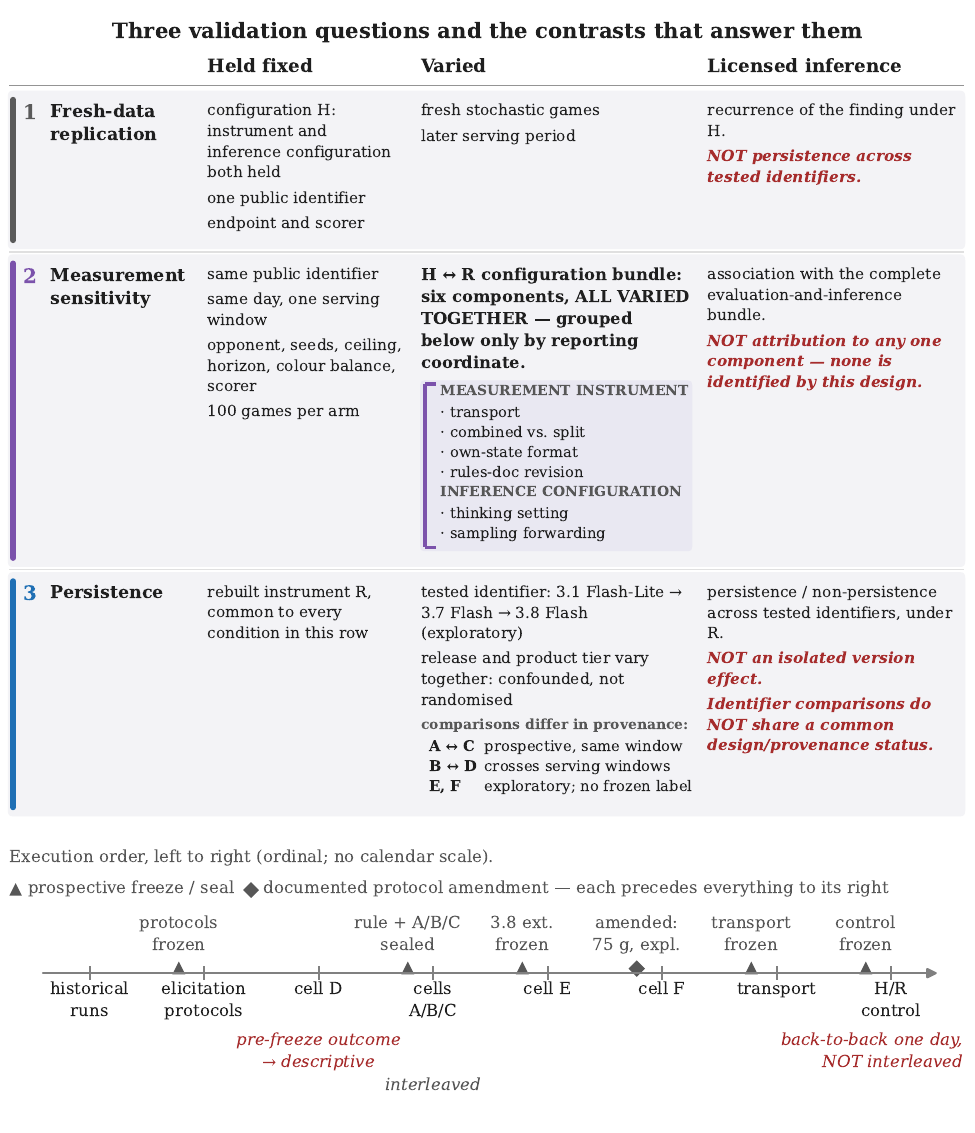}
\caption{\textbf{Three validation questions and the contrasts
that answer them.} Row~2's \emph{Varied} cell shows the H/R bundle as a
single treatment: all six components varied together, bracketed as one unit
and grouped below only by reporting coordinate (four measurement-instrument,
two inference-configuration), so the contrast spans both coordinates of the
scheme in Section~\ref{sec:discussion}. That grouping is operational, not a
decomposition: neither a component nor either sub-bundle is identified.
Row~3 records that identifier comparisons do not share a common
design/provenance status. Cell~D's outcome preceded the rule freeze, so its
label is descriptive.}
\label{fig:design}
\end{figure}

\paragraph{Statistical and provenance conventions.} Endpoints, cutoffs, and
the classification rule were frozen before the conditions they judge, and
every condition carries an explicit status: confirmatory where it was
prospectively frozen, descriptive where its outcome preceded the relevant
freeze, and exploratory where it was downgraded on operational grounds. All
confidence intervals are game-clustered bootstrap intervals, because many
named-candidate events occur within a single game and are not independent;
resampling is therefore at the level of whole games, with the number of
replicates and the seed fixed in advance. Analyses whose specification
postdates the results they describe --- the depth and state-richness
stratifications, and the actionability decomposition --- are labelled as
such at every point of use, and cross-cell contrasts are exploratory
relative to the frozen design, which ran no cross-cell test. Where a
stratified analysis has fewer than 100 contributing game clusters, it is
reported descriptively rather than inferentially under the study's existing
support convention. Finally, every reported quantity was recomputed by
independent reimplementations that share no scoring code with the canonical
pipeline and were built from the frozen endpoint definitions alone; all
quantities reported in this paper derive from a frozen, independently
audited result set. Per-condition game counts, identifiers, freeze
sequence, and the integrity record are in Appendices~\ref{app:hrcontrol},
\ref{app:provenance} and~\ref{app:independent}.
Figure~\ref{fig:design} summarises the three questions, what each contrast
holds fixed and varies, what each therefore licenses, and the execution and
freeze order that the following sections rely on.

\section{Replication and Measurement Sensitivity}
\label{sec:replication}

This section answers two of the three questions, and keeps them apart.
The first asks whether the historical finding recurs on newly generated
games when its own configuration is rerun. The second asks how much the
endpoint moves when the configuration is rebuilt with the model identifier
and the serving day held fixed. Persistence across identifiers is a
separate question and is deferred to Section~\ref{sec:persistence}.

\paragraph{Fresh-data replication under the historical configuration.}
Rerunning H in a later serving period, on newly generated games, recovers
the historical result. The historical 60-game arm --- the reference arm for
this comparison, because it is the larger of the two historical runs and
the only one this contrast uses --- gives
$\Delta_{R1} = +0.0696$, 95\% CI $[+0.0428, +0.0937]$. The rerun
(``H-current'', 100 games, 2{,}444 named-candidate events across 94
scoreable games) gives $+0.0530$, $[+0.0329, +0.0714]$. Both intervals lie
entirely above zero, so both receive the frozen endpoint-status label
\textsc{replicates}: the model scores worse than its matched comparator,
in the direction the historical degradation had. The label is doing only
that work. The study-design claim is the separate one, and it is stronger
than the label: new games were generated under the historical
configuration, so this is a \emph{fresh-data replication under H}. The
20-game historical arm ($+0.0593$, $[+0.0292, +0.0931]$) is reported as
descriptive sensitivity only and is never pooled with the 60-game arm.

The contrast between them --- H-current minus the historical 60-game arm
--- is $-0.0166$, 95\% CI $[-0.0483, +0.0157]$, which spans zero. This is
the comparison that would speak to a serving-period contribution, and it
is inconclusive. The inconclusiveness is a statement about precision
rather than about the effect: the historical arm is fixed at 60 games, so
the interval's half-width cannot fall below $0.0254$ however much new data
is collected, and the realised half-width is $0.0320$. Reading this
interval as evidence of equivalence, or as evidence of no serving-period
effect, would be a mistake in both directions.

Format-level behaviour was not identical across the two windows even under
H, and reporting it is part of reporting the replication honestly.
Well-formed elicitation ran at 82.3\% in the archived historical arm
against 72.1\% (1{,}212 of 1{,}682) in H-current, a difference of $-10.3$
percentage points, inside the prospectively frozen $\pm 15$-point
diagnostic band. That records a difference in format-level behaviour across
windows under one configuration. It is not evidence of a mechanism, and not
evidence of a backend change.

The reconstructed Gemini 3.1 conditions under R, reported in
Section~\ref{sec:persistence}, recover the same degradation direction but
at materially attenuated magnitude --- roughly one-third to one-half that
of the historical references, at both the 4K and 16K ceilings. That
attenuation is descriptive. The reconstructed and historical conditions
differ in measurement configuration and serving period at once, so it
isolates no configuration component, no serving-period contribution, and
no other cause; it is the reason for the control described next.

\paragraph{Measurement sensitivity: H-current versus R-current.} Both arms
ran in the same serving window, under the same public model identifier,
with output ceiling, opponent, game horizon, seed schedule, colour balance,
and scorer held fixed, at 100 games each (Section~\ref{sec:design}). Under
H the deficit is $+0.0530$, $[+0.0329, +0.0714]$ (\textsc{replicates}).
Under R, in the same window, it is $+0.0101$, $[-0.0036, +0.0234]$, whose
interval spans zero (\textsc{unresolved}). The H$-$R contrast is
$\Delta = +0.0429$, 95\% CI $[+0.0182, +0.0667]$, entirely above zero. The
shared environment-seed schedule equalises the opponent's randomisation
across arms but does not produce matched games, so the two arms are
analysed unpaired, with resampling clustered on whole games.
Integrity was complete on both arms: 100 of 100 games on the frozen seed
schedule, no errored games, no game reaching the 150-ply horizon, no
truncated generations, and every configuration field single-valued within
its arm.

Well-formedness was also not identical across the two configurations, and
because responses that are not well formed are excluded rather than mended,
each arm is scored on its own eligible population. A
configuration-dependent change in which outputs are well formed enough to
enter the endpoint is itself part of the observed measurement sensitivity;
consequently the H$-$R contrast characterises the complete measurement
endpoint, including the scoreable population that endpoint induces, rather
than belief quality on a fixed common subset.

What the control establishes is a \emph{material endpoint difference
associated with the bundled evaluation and inference configuration},
measured under one public identifier on one day, subject in full to the
design qualifications set out in Section~\ref{sec:design}. Two
consequences of those qualifications matter for reading the number. The
difference is an association with the bundle, not a demonstration that the
bundle produced the historical result: we do not claim it explains the
historical gap, and the historical-versus-reconstructed difference itself
remains jointly indexed by configuration and serving period. And the
comparison speaks to this identifier on this day, not to configuration
sensitivity in general.

\begin{figure}[p]
\centering
\includegraphics[width=\linewidth]{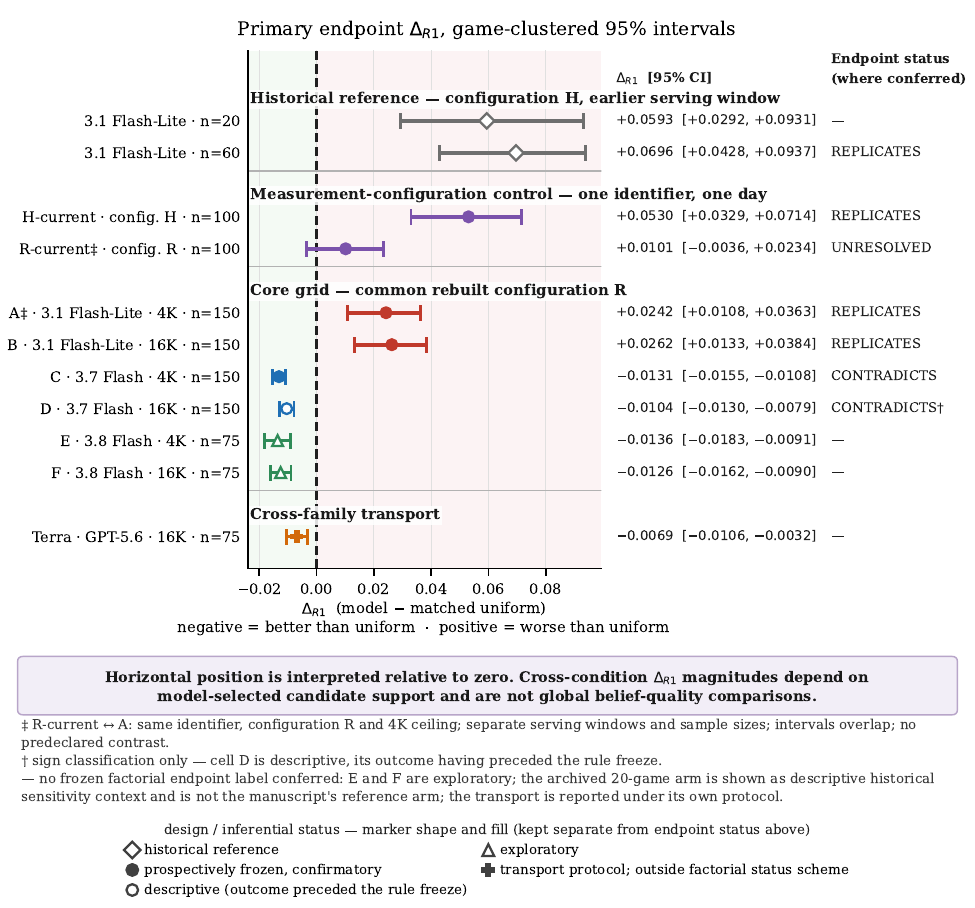}
\caption{\textbf{Primary endpoint $\Delta_{R1}$ across every condition
in the study.} Intervals are game-clustered 95\%
bootstrap intervals at the canonical board-ply $\geq 8$ cutoff.
\emph{Two different kinds of status are shown on two separate channels and
must not be conflated:} the marker gives each row's design and inferential
status, the right-hand column the frozen endpoint status where the design
confers one. Groups are distinct study designs that
share an endpoint scale; co-location does not license arbitrary cross-group
contrasts. Aggregate R1 is support-dependent: scalar $\Delta_{R1}$
magnitudes are conditional on model-selected support and are not global
belief-quality comparisons, so the plot reads by sign and interval position
relative to zero and by endpoint-status pattern, not by scalar distance.
$\ddagger$~R-current and cell A are distinct executions under the same
identifier, configuration R and 4K ceiling, run in separate serving windows
at different sample sizes; their intervals overlap and no predeclared
contrast relates them --- the marker explains why two closely related
executions carry different frozen labels, and implies neither a paired nor
a controlled comparison, nor a serving-period or sample-size explanation.
$\dagger$~Cell D's \textsc{contradicts} is a sign classification only; the
cell is descriptive, its outcome having preceded the rule freeze. Cells E
and F are exploratory and carry no factorial endpoint label. The
cross-family transport is a separate protocol reported outside the
factorial status scheme. The archived 20-game historical arm is shown as
descriptive historical sensitivity context and is not the historical
reference arm.}
\label{fig:forest}
\end{figure}

\paragraph{R-current beside cell A.} Under R the deficit is directionally
positive both in this control arm ($+0.0101$, $[-0.0036, +0.0234]$, 97
scoreable games of 100) and in cell A of the grid below, which shares the
tested identifier, the rebuilt configuration R and the 4K ceiling
($+0.0242$, $[+0.0108, +0.0363]$, 145 scoreable games of 150). The control
arm's interval spans zero and is labelled \textsc{unresolved}; cell A's
does not and is labelled \textsc{replicates}. The two intervals overlap,
the point estimates differ, and the two executions ran in different serving
windows at different sample sizes under no predeclared
contrast. The differing frozen labels are therefore not evidence of a
serving-period effect, of a sample-size effect, or of any other isolated
cause; they are two separate executions whose classification rule happened
to fall on opposite sides of zero. Figure~\ref{fig:forest} places every
condition in the study on the one primary-endpoint scale, with each row's
design status and, where the design confers one, its frozen endpoint status.

Read together, the two halves of this section already make the paper's
narrower methodological point. The same nominal model, measured in the same
window through two configurations of the same nominal evaluation, produces
endpoints one of which is labelled \textsc{replicates} and one of which is
\textsc{unresolved}. Whatever else is true of the finding, a report of it
that names only the model is not a report of a reproducible measurement.

\section{Persistence Under the Rebuilt Instrument}
\label{sec:persistence}

The persistence question is asked under a single common instrument, so that
a difference between identifiers cannot be a difference between harnesses.
All six cells below were evaluated under the rebuilt configuration R.

\paragraph{The core grid.} Table~\ref{tab:core} reports all six cells. The
originally tested identifier retains the deficit at both ceilings: cell A
(Gemini 3.1 Flash-Lite, 4K) gives $+0.0242$, $[+0.0108, +0.0363]$ over
6{,}611 events in 145 scoreable games, and cell B (16K) gives $+0.0262$,
$[+0.0133, +0.0384]$ over 6{,}687 events in 147 games. Both are
\textsc{replicates}. The next identifier tested reverses the endpoint
direction at both ceilings: cell C (Gemini 3.7 Flash, 4K) gives
$-0.0131$, $[-0.0155, -0.0108]$ over 21{,}550 events in 150 games, and
cell D (16K) gives $-0.0104$, $[-0.0130, -0.0079]$ over 21{,}498 events in
150 games. Both are \textsc{contradicts}, and the sign classification is
therefore unchanged across the two tested ceilings within each identifier.

\begin{table}[t]
\centering
\caption{\textbf{Core grid, all six cells, under the common rebuilt
configuration.} R1 is the primary endpoint; positive means worse than the
matched comparator, negative better. Classification is the frozen
interval-versus-zero rule of Section~\ref{sec:measurement}, and is an
endpoint-status label. Cell D's label is descriptive: its outcome preceded
the freeze of that rule. Cells E and F are exploratory and carry no
classification. The cross-family transport is reported in the text, not
here, because it sits outside this design and its classification scheme.}
\label{tab:core}
\small
\begin{tabular}{@{}llrrlr@{}}
\toprule
Cell & Model & Ceiling & Games & R1 $\Delta$ [95\% CI] & Status \\
\midrule
A & Gemini 3.1 Flash-Lite & 4K  & 150 & $+0.0242$ $[+0.0108, +0.0363]$ & \textsc{replicates} \\
B & Gemini 3.1 Flash-Lite & 16K & 150 & $+0.0262$ $[+0.0133, +0.0384]$ & \textsc{replicates} \\
C & Gemini 3.7 Flash      & 4K  & 150 & $-0.0131$ $[-0.0155, -0.0108]$ & \textsc{contradicts} \\
D & Gemini 3.7 Flash      & 16K & 150 & $-0.0104$ $[-0.0130, -0.0079]$ & \textsc{contradicts}$^{\dagger}$ \\
E & Gemini 3.8 Flash      & 4K  & 75  & $-0.0136$ $[-0.0183, -0.0091]$ & exploratory \\
F & Gemini 3.8 Flash      & 16K & 75  & $-0.0126$ $[-0.0162, -0.0090]$ & exploratory \\
\bottomrule
\end{tabular}

\vspace{0.3em}
\begin{minipage}{0.93\linewidth}\footnotesize $^{\dagger}$Descriptive only.
Cell D ran in the immediately preceding serving window and completed before
the A/B/C interleaved window opened; its outcome therefore preceded the
rule freeze and it is not prospectively confirmatory.\end{minipage}
\end{table}

Not all four cells support the same inference, and the asymmetry is a
property of when they ran rather than of what they measured.
\textbf{A versus C is the strongest cross-identifier comparison available}:
one common output ceiling, one common configuration, both cells
prospectively frozen, and interleaved execution within one serving window.
\textbf{B versus D is interpretively weaker} on one specific ground --- D
ran in the immediately preceding serving window and completed before the
A/B/C interleaved window opened, so any B-versus-D contrast crosses
serving windows, which is exactly the confound the same-window control of
Section~\ref{sec:replication} exists to bound. For completeness the
cross-identifier contrasts are $+0.0373$, $[+0.0233, +0.0498]$ at 4K
(A versus C) and $+0.0367$, $[+0.0236, +0.0492]$ at 16K (B versus D); both
are exploratory relative to the frozen design, which specified no
cross-cell test. And whichever pairing is used, the two identifiers differ
in release \emph{and} in product tier, neither of which this design
randomises, so what the grid contrasts is tested identifiers.

\paragraph{Exploratory extension to a third identifier.} Cells E and F
extend the design, under the same configuration R, to Gemini 3.8 Flash ---
an identifier released during execution of the core grid. Both fall on the
same side of the comparator as the Gemini 3.7 cells, at both ceilings
($-0.0136$ at 4K, $-0.0126$ at 16K), so the reversal is not confined to a
single tested release. These two cells were downgraded to exploratory on
operational grounds before either was scored, as described in
Section~\ref{sec:design}, and they carry no confirmatory weight anywhere in
this paper: they show a pattern is not isolated, and nothing stronger.

\paragraph{What the two ceilings do and do not establish.} Belief
generation did not bind the tested ceilings in any core cell: belief-call
truncation was zero in all four of A, B, C and D. The grid therefore
establishes that the sign classification is \emph{invariant over the tested
non-binding ceilings} --- which is a statement about robustness of the
endpoint to a parameter that never became active, not an estimate of a
token-budget effect. No isolated token-budget effect is claimed or
estimable here. One exploratory cell does speak to the question from the
other side: in cell E the 4K ceiling genuinely bound, truncating 24.3\%
(384 of 1{,}580) of belief calls, and the cell still landed on the negative
side of the comparator. That weighs against a pure truncation account of
the reversal without estimating a causal effect of truncation, and cell F
at 16K, where nothing truncated, lands in the same place.

\paragraph{Cross-family transport.} The reversal is not confined to one
provider lineage. A prospectively frozen cross-family transport ---
GPT-5.6, accessed as the transport identifier ``Terra'' over a different
provider API, at the 16K ceiling, with the same fixed opponent, the same
elicitation architecture and the same 75 environment seeds as cell F ---
gives $\Delta_{R1} = -0.0069$, 95\% CI $[-0.0106, -0.0032]$ over 9{,}873
events in 75 games, on the opposite side of the comparator from the
historical degradation. Two things about its status are deliberate. It is
reported outside the factorial and outside the factorial's classification
scheme, which is reserved for the Gemini reconstruction design, so it
carries the protocol's own descriptive label rather than
\textsc{replicates} or \textsc{contradicts}. And it is not simply another R
cell: it preserves R's elicitation architecture over a provider-specific
implementation with its own request shaping and reasoning configuration,
which makes it transport-analogous to R rather than transport-identical.
Its sample size was fixed prospectively rather than by an outcome-dependent
stopping rule. Terra is the only cross-family transport tested, and its
difference attenuates under stricter depth restrictions
(Section~\ref{sec:comparator}).

\section{Behavioural Decomposition}
\label{sec:decomposition}

\begin{figure}[t]
\centering
\includegraphics[width=0.99\linewidth]{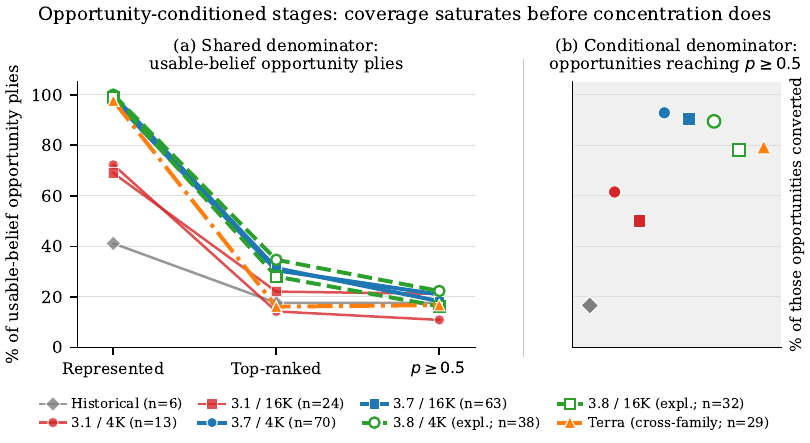}
\caption{\textbf{Opportunity-conditioned stages, split at the denominator
change.} \textbf{(a)} Represented (named at all), top-ranked, and
$p\geq0.5$ share one denominator --- usable-belief opportunity plies ---
so the three stages are connected. \textbf{(b)} Conversion is conditional
on the opportunities that reached $p\geq0.5$, a different and much smaller
population, so each condition's $n$ is given in the legend and no line
joins it to (a). Circles mark the 4K and squares the 16K output ceiling;
the exploratory Gemini 3.8 series are hollow and dashed; the cross-family
transport is the triangle. All values are read from the corrected
opportunity artifact. The figure is descriptive and identifies no
mechanism.}
\label{fig:opportunity}
\end{figure}

A secondary analysis conditions on actionability, which neither R1 nor R2
does. An \emph{opportunity} exists on one of the model's own turns when the
opponent's true Regent is among its legal capture targets in the exact
pre-move position; capturing it wins immediately, so both the availability
of the action and its relevance are unambiguous.
Figure~\ref{fig:opportunity} decomposes belief quality on that population
into four stages for all six cells, the historical corpus, and the
cross-family transport. In the newer conditions (C through F) the true
hidden state is \emph{represented} --- named as a candidate at all --- on
97.9--100\% of usable-belief opportunity plies, but is \emph{top-ranked} on
only 28.1--34.7\% of them and assigned $p \geq 0.5$ on only 16.3--22.4\%.
Once that confidence threshold is reached the available win is usually
taken: 65 of 70 such opportunities in cell C, 57 of 63 in D, 34 of 38 in E,
25 of 32 in F, and 23 of 29 under the cross-family transport. The
historical corpus sits far below on the first stage, with the true state
represented on 41.2\% of its usable-belief opportunity plies, top-ranked on
17.6\%, and at $p\geq0.5$ on 17.6\%, converting 1 of 6. Coverage of the
hidden state, in short, saturates before concentration on it does.

Three limits bound what that pattern can mean. First, it identifies no
mechanism. The staged gap between representation and confidence co-varies
with tested identifier, and this analysis does not establish why: it is
compatible with better hidden-state tracking, with more selective
assignment of high confidence, or with both, and nothing here separates
them. We therefore do not claim that the later identifiers learned to track
the hidden state, that tracking causally improved, or that tracking has
been distinguished from confidence calibration. Second, the analysis is
ancillary and descriptive throughout --- it introduces no inferential test,
does not modify R1 or R2, and its provenance differs by condition: the
historical corpus and cells A--D were scored post-hoc under a
specification written after those results existed, the E/F specification
was frozen before scientific unseal though not prospective in the strong
sense the core endpoints were, and the transport's opportunity analysis was
prospectively frozen before execution. Third, the population is
opponent-dependent. Opportunity density is a property of the fixed
opponent's play, so these rates characterise that regime rather than an
opponent-invariant property of a model, and the high whole-game win rates
in the newer conditions (98.7--100\%, and 89.3\% for the transport) are
consistent with the staged gap rather than in tension with it: a game needs
one terminal capture, while a trajectory can contain many actionable
opportunities. Full cross-tabulations, denominators, and the en-passant
implementation correction that fixes the opportunity denominators are in
Appendix~\ref{app:opportunity}.

\section{Comparator and Robustness Context}
\label{sec:comparator}

Four things materially constrain how the endpoints above may be read, and
this section states them compactly rather than re-deriving them.

\paragraph{Support dependence, and what R2 adds.} R1's aggregate is
support-dependent because the model selects its own event set
(Section~\ref{sec:measurement}), and the tested identifiers differ
materially on exactly that axis: 4.50 and 4.48 named candidates per scored
ply in cells A and B, against 8.39 and 8.68 in C and D, and 8.32 for the
cross-family transport measured over its eligible belief plies.
R2, the complementary endpoint, scores the full elicited partition
including the residual, so it also penalises leaving the true state
unnamed: $+0.0575$ $[-0.0413, +0.1509]$ in A, $+0.1473$
$[+0.0493, +0.2392]$ in B, $-0.1199$ $[-0.1416, -0.0979]$ in C, $-0.1028$
$[-0.1269, -0.0783]$ in D, and $-0.1307$ and $-0.1233$ in the exploratory
E and F. R2 moves with R1 in three of the four core cells; cell A's
canonical interval includes zero, and we report that rather than reading
the point estimate alone. R2 magnitudes are support-dependent in the same
way as R1's. The residual itself is where much of the difference sits: the
true state falls in the unnamed residual on 23.0\% and 30.8\% of scored
plies in A and B, against 1.5\% and 1.1\% in C and D, and 1.77\% for the
transport over its diagnostics population.

\paragraph{Depth and state richness.} Both identifier groups were re-scored
at two stricter prespecified depth cutoffs. The Gemini 3.1 cells strengthen
monotonically with depth (A rises to $+0.0411$ at ply $\geq 10$ and
$+0.0618$ at ply $\geq 14$; B to $+0.0442$ and $+0.0703$), while every
Gemini 3.7 and 3.8 condition shrinks toward zero, with cell C the only one
retaining an interval entirely below zero at ply $\geq 14$ ($-0.0053$,
$[-0.0077, -0.0029]$) and cell D crossing into \textsc{unresolved} there.
The transport shows the same shrinking signature, approximately zero by
ply $\geq 14$. This is a descriptive re-analysis, not a preregistered
interaction test, and no cutoff revises any condition's canonical
classification. A separate exploratory stratification asks whether the
reversal is an artifact of late-game candidate-set collapse by restricting
scoring to plies with at least eight live opponent pieces --- over 80\% of
eligible events in each core cell. It is not: the cross-identifier
difference persists there at both ceilings, $+0.0277$
$[+0.0128, +0.0414]$ at 4K and $+0.0310$ $[+0.0172, +0.0446]$ at 16K,
modestly attenuated relative to the all-support contrasts
(Appendix~\ref{app:kstrat}). The two exploratory cells fall
below the study's 100-cluster support floor in that stratum and are
reported descriptively, outside the inferential claim.

\paragraph{The capture-time diagnostic.} A separate diagnostic scores only
the captured-square event, as the ratio of model to matched-uniform Brier
loss. It falls from roughly $4.2$--$4.6\times$ the comparator for
Gemini 3.1 Flash-Lite to approximately parity for Gemini 3.7 Flash
($0.97$--$0.99\times$). This is a different population from the primary
endpoints and from the transport's own capture diagnostic, and the three
should not be placed on one axis. Improvement to parity is not resolution:
at the 16K ceiling the absolute R2 loss is still $0.7375$ against the
matched comparator's $0.8403$, so substantial absolute error persists in
the improved conditions.

\paragraph{How weak is the matched comparator?} A uniform prior over live
opponent pieces is a deliberately undemanding benchmark, and it is fair to
ask whether beating it or losing to it means much. One piece of context
bounds the answer on this population. A predecessor study built an
auxiliary tracker that updates a distribution over which opponent piece
holds royal status using only deductively valid rules keyed to publicly
observable events (Appendix~\ref{app:pubtracker}). Scored against matched
uniform on the historical capture-only population at board ply $\geq 8$, it
showed only a small advantage in one historical batch ($n=77$) and no clear
advantage in the other ($n=220$). We therefore use it only as
limited-headroom context, and not as a stronger comparator, not as a
tractability certificate, and not as evidence that the hidden state is
predictable after a shift. Its own scope is narrow on three axes at once
--- one tracker design, one opponent, one historical population --- and the
opponent's shift timing is only partly state-responsive by policy
(Section~\ref{sec:design}), which bounds what any public-information
predictor could achieve here.

\section{Discussion and Scope}
\label{sec:discussion}

The results support a narrow claim precisely, and the boundaries below are
the conditions under which that claim holds rather than a list of
shortcomings. Stating them as scope is the honest framing: each one is a
choice the design made, and most of them could only be relaxed by a
different study.

What the results establish is that three validation questions about one
behavioural finding can dissociate. Rerunning the historical measurement
configuration on newly generated games recovers the finding. Substituting a
rebuilt configuration within one day and under one public identifier moves
the endpoint materially. And under that common rebuilt configuration, the
finding does not carry over to the next identifiers tested. No one of those
three statements implies either of the others, which is the point.

\paragraph{Scope conditions.} The boundaries below group into four kinds.

\begin{itemize}\itemsep2pt \parskip0pt
\item \textbf{Environment and opponent.} All results come from one
controlled sequential hidden-state environment, chosen because it records
latent truth exactly, and every interactive condition faced one fixed
Tier-1 heuristic opponent. Trajectory distributions, opportunity density,
and how far the hidden state is predictable from public play are therefore
properties of that regime rather than of the models; whether the
dissociation appears in other action-time settings is untested here.
\item \textbf{Identifier and provider.} The tested Gemini identifiers
differ in release \emph{and} product tier, neither of which is randomised,
so every cross-identifier statement is about tested identifiers rather
than an isolated version effect (Section~\ref{sec:persistence}). The later
identifiers were evaluated under R only, so persistence here means
persistence under the common rebuilt instrument, not across instruments.
One cross-family transport shows the reversal is not confined to a single
provider lineage, which is not provider-level generality.
\item \textbf{H/R design.} The configuration control varies a
six-component bundle in one same-day, back-to-back comparison
(Section~\ref{sec:design}). It bounds the configuration question as a
whole and resolves no component within it, and the serving-period
contribution remains unresolved at a precision the fixed historical sample
size bounds from below --- neither an effect nor an equivalence.
\item \textbf{Measurement and provenance.} R1's aggregate remains
support-dependent because the model selects its own event set
(Section~\ref{sec:measurement}), so the confirmatory design rests on sign
and classification rather than magnitude. And no mechanism is identified:
nothing here explains why the later identifiers differ --- not
serving-side change, not the output ceiling, and not the gap between
representing the hidden state and concentrating probability on it.
\end{itemize}

\paragraph{An asymmetry in endpoint-identity verification.} One provenance
gap deserves stating in its own right. Only the Gemini 3.8 identifier
received an explicit provider metadata-resolution audit, in which the
provider-returned resolved version string was recorded and carried no
preview or experimental marker. No equivalent resolved-version audit exists
for the other two Gemini identifiers in any execution window, including the
historical runs and their reconstruction. This is an asymmetry in how
firmly the served endpoint behind each identifier was verified, not
evidence that any identifier resolved to something unexpected: nothing here
indicates that the unaudited identifiers were wrong, only that we cannot
demonstrate from our own records what they resolved to. It is also exactly
the condition under which this paper's own argument applies to itself ---
an identifier without a resolution record is a weaker provenance claim than
one with it.

\paragraph{Why the three questions should not be collapsed.} They differ in
what is held fixed and therefore in what a result can license.
\emph{Replication} holds the measurement configuration fixed and varies the
data, answering whether the finding recurs. \emph{Measurement sensitivity}
holds the identifier and serving period fixed and varies the
configuration, answering whether the finding is a property of the
measurement. \emph{Persistence} holds the configuration fixed and varies
the tested identifier, answering whether the finding still describes what
is being served now. Reporting a single rerun answers at most one of them
while appearing to answer all three: a successful rerun under the original
harness says nothing about the next release, and a failed rerun under a
rebuilt harness cannot distinguish a changed model from a changed
instrument. That is not a hypothetical concern in this study. The same
identifier, in the same window, measured through two configurations,
produced one endpoint labelled \textsc{replicates} and one labelled
\textsc{unresolved}.

\paragraph{Methodological implication.} A hosted-model behavioural claim
should be indexed to four things, not one: the \emph{tested identifier},
the \emph{serving period} in which it was measured, the \emph{measurement
instrument} that elicited and scored the behaviour, and the
\emph{inference configuration} under which the model was run. A claim
indexed only to a model name is not wrong so much as incomplete --- it does
not say enough for anyone, including its own authors, to know what a
subsequent rerun is comparing itself to. The design artefacts here are
reusable in a way a single measured outcome is not: a frozen elicitation
architecture, frozen endpoints, and a frozen actionability definition can
be applied to any future release at known marginal cost, whereas the
measured outcome may already describe a service that is no longer being
served.

\section{Related Work}
\label{sec:related}

\textbf{Calibration and probabilistic belief elicitation.} Work on
confidence and calibration in language models has largely evaluated
probabilities in static settings, where the model answers a question or
labels an instance and reports confidence in that same act
\citep{kadavath2022know,kim2026same,li2025conftuner,wang2026calibrating,
leng2025taming,ffrenchconstant2026confidencebench}. That literature
establishes the phenomena we rely on --- that calibration is often poor,
that overconfidence is systematic, and that calibration quality varies
sharply across families and versions --- and one such benchmark
independently places the model at the centre of this study at the poor end
of its own calibration distribution on an unrelated task family
\citep{ffrenchconstant2026confidencebench}. What that design cannot
separate is the scored answer from the stated confidence, because they are
the same output. We move the measurement into a sequential interactive
setting and elicit belief about an already-determined hidden state in a
call separate from the one that chooses the action, which makes belief
quality and action quality separately scoreable.

\textbf{Interactive and hidden-state evaluation.} Competitive-game
platforms \citep{olszewska2025kaggle} and agent benchmarks under partial
observability \citep{singh2026agentbrace,samanta2026bayesbench}, together
with social-deduction settings built to probe hidden-role inference
\citep{agarwal2025wolf,karpov2026mafiascope}, increasingly place models
where latent state must be inferred from interaction history. Most score
task success or action quality; where belief is elicited at all, it is
typically optimised against downstream reward rather than scored against
the action it informed. The environment used here contributes exactly
recoverable hidden-state ground truth together with explicit probability
elicitation, which is what permits a proper score at the moment of action.

\textbf{Hosted-endpoint temporal and configuration variability.} A
longitudinal study tracking GPT models on materials-science tasks over 18
months reports 9--43\% performance variation, attributes part of it to
silent endpoint updates and version swaps, and recommends recording exact
version strings and evaluation dates \citep{venugopal2026probing}. A
supply-side measurement study makes a structurally related point from the
provider side: a hosted model identifier names a provider-specific,
time-varying service rather than a fixed artefact \citep{li2026same}.
Separately, static factuality benchmarks age as the facts they test become
outdated \citep{jiang2026benchmarks} --- the converse mechanism, in which
benchmark content drifts relative to the world rather than the service
drifting relative to a fixed evaluation. Our contribution here is the
instrument dimension these accounts do not isolate: reconstructing the
measurement configuration is itself a source of measurement difference,
and it is separable from serving period by design rather than by argument.

\textbf{Reproducibility and replication methodology.} We use the standard
distinction between recovering reported results from archived artifacts and
conducting a fresh study on newly generated data
\citep{nasem2019reproducibility}, and we keep it visible in the results
themselves rather than confining it to terminology: the historical rerun is
a fresh-data replication, while the factorial cells are evaluations under a
rebuilt configuration and are never described as replications of the
historical experiment.

\textbf{What is and is not new here.} We do not claim to discover that
hosted APIs vary over time, nor that measurement details matter; both are
established, and the studies above say so. The contribution is evidential
and specific: using action-time proper scoring against exactly recorded
hidden-state ground truth, we show that fresh-data replication,
across-identifier persistence, and measurement-configuration sensitivity
can \emph{dissociate within a single behavioural evaluation} --- the same
finding replicating under its own instrument, moving materially when the
instrument is rebuilt under one identifier on one day, and failing to
persist across the next identifiers tested under that common instrument.
The accompanying elicitation analysis is a second, smaller contribution of
the same kind: it shows that the format in which belief is elicited is not
a neutral implementation choice, measured against prospectively frozen
decision rules rather than asserted.

\section{Conclusion}
\label{sec:conclusion}

Rerun on newly generated games under its own historical measurement
configuration, the previously reported action-time belief deficit recovers:
the endpoint remains on the same side of its matched comparator, with the
frozen classification unchanged. Substituting a rebuilt measurement
configuration in the same serving window and under the same public model
identifier moves that endpoint materially, by an amount comparable to the
deficit itself, and the design associates the shift with the configuration
bundle as a whole rather than with any component of it. Under that common
rebuilt configuration the deficit then fails to persist: the next Gemini
identifier tested reverses the endpoint direction at both tested output
ceilings, as does a further exploratory release and a cross-family
transport.

The behavioural decomposition adds a specific shape to the improvement
rather than a mechanism for it. On states where the hidden target was
legally actionable, the newer conditions represent the true state almost
whenever they can, while ranking it first or assigning it majority
probability far less often --- coverage saturates before concentration
does. That pattern is compatible with better tracking, with more selective
confidence, or with both, and this study does not separate them.

What follows for practice is a reporting discipline rather than a
correction to any particular result. A behavioural claim about a hosted
model should be indexed to the tested identifier, the serving period, the
measurement instrument, and the inference configuration, because a rerun
that shares only the model name is not necessarily measuring the same
object. Reversal is also not resolution: substantial absolute error
persists in the improved conditions, and nothing here establishes why the
transition occurred or whether it will hold in the next release tested.

%


\bibliography{references}
\bibliographystyle{tmlr}

\appendix


\section{Appendix}
\label{app:appendix}

This appendix provides the full detail supporting the main text. No claim
in Sections~\ref{sec:intro}--\ref{sec:conclusion} depends on the reader
consulting it. It is written to be audited rather than read
continuously: definitions are given exactly, every table carries its own
population and status, and interpretation is confined to short notes
attached to the quantity being interpreted. All reported quantities derive
from a frozen, independently audited result set.

\subsection{Full environment rules}
\label{app:rules}

Condensed from the predecessor study that introduced the environment
\citep{joshi2026confident}; platform, interface and data-consent
provisions that do not affect the reported game mechanics are omitted.

\textbf{Core principle.} A chess variant in which royal status can be
secretly transferred between a player's own pieces. The objective is to
capture the opponent's \emph{Regent} --- their current royal piece ---
which may or may not be their Original King.

\textbf{Terminology.} \emph{Regent}: the piece currently holding royal
status; capturing the opponent's Regent wins immediately. \emph{Original
King}: the piece that started the game as king. \emph{Crown Shift}: the
secret act of transferring royal status to another piece. \emph{King
Mode}: the Original King is the Regent, and standard chess rules apply to
it. \emph{Regent Mode}: a different piece is Regent and gains royal
immunity.

\textbf{Setup.} Both players begin in King Mode with their Original King
as Regent; standard chess setup and rules otherwise apply.

\textbf{Crown Shift.} Available to each player from the start of their 4th
turn, and again 15 completed moves after each use, tracked independently
per player. Usable even while in check or checkmate. On one's own turn,
without declaration, a player may secretly designate any of their own
pieces --- including the Original King, returning to King Mode --- as the
new Regent. The shift does not consume the turn: a regular move must still
follow. If checkmated in King Mode with Crown Shift available, the player
may shift before any other action; shifting to a non-king piece nullifies
the checkmate, while shifting to the king, or not shifting, ends the game
in a loss. A shift onto an attacked square risks immediate capture and
loss --- the crown may be placed anywhere, including under attack.

\textbf{Game modes.} \emph{King Mode}: standard chess rules and check
restrictions apply to the Original King; castling is permitted, and
possible, only here. \emph{Regent Mode}: the Regent moves by its own piece
type's rules, is immune to check restrictions, and its capture ends the
game immediately; the demoted Original King retains one-square movement,
may sit on or move into attacked squares, cannot castle, and its capture
is ordinary material loss only.

\textbf{Win conditions.} Capturing the opponent's Regent wins immediately.
In King Mode, checkmate on the Original King with no Crown Shift escape is
a loss. In Regent Mode, traditional checkmate cannot occur, because the
Regent is immune to check.

\textbf{Other rules.} A promoted Regent pawn automatically transfers royal
status to the promoted piece. Check is always announced against the
Original King regardless of mode, and a Regent-Mode player may legally
ignore it. Traditional stalemate is abolished: a player with no legal move
passes automatically, with a draw only if both players pass on consecutive
turns. A pass does not count toward the Crown-Shift cooldown or the
50-move no-progress rule, but does count toward the 75-move backstop.
Draws also follow from the 50-move rule (reset by capture or shift); the
75-move backstop (not reset by shift, counted in total plies, to prevent
indefinite stalling); threefold repetition on visible board state only, so
a hidden Crown Shift between occurrences does not prevent the draw; and
the bare-kings dead position, all other material configurations remaining
live because even a lone knight can win by capturing the Regent. Castling
requires King Mode, the Original King and a non-Regent rook. All other
standard chess rules, including en passant, remain in effect.

\subsection{Depth-sensitivity and support diagnostics}
\label{app:depth}

\textbf{Depth cutoffs.} The canonical cutoff is board ply $\geq 8$. Two
stricter prespecified cutoffs are reported for every condition on both
endpoints. No cutoff revises any condition's canonical classification,
which is the frozen basis for every main-text claim. Classifications are
shown only where the design confers them: cells A--C prospectively, cell D
descriptively, and E/F not at all.

\begin{center}
\scriptsize
\begin{tabular}{@{}llll@{}}
\toprule
Cell & ply $\geq 8$ & ply $\geq 10$ & ply $\geq 14$ \\
\midrule
\multicolumn{4}{@{}l}{\emph{R1 $\Delta$ [95\% CI], classification}}\\
A & $+0.0242$ [$+0.0108,+0.0363$] \textsc{repl.} & $+0.0411$ [$+0.0280,+0.0531$] \textsc{repl.} & $+0.0618$ [$+0.0489,+0.0737$] \textsc{repl.} \\
B & $+0.0262$ [$+0.0133,+0.0384$] \textsc{repl.} & $+0.0442$ [$+0.0316,+0.0563$] \textsc{repl.} & $+0.0703$ [$+0.0576,+0.0825$] \textsc{repl.} \\
C & $-0.0131$ [$-0.0155,-0.0108$] \textsc{contr.} & $-0.0094$ [$-0.0118,-0.0069$] \textsc{contr.} & $-0.0053$ [$-0.0077,-0.0029$] \textsc{contr.} \\
D$^{\dagger}$ & $-0.0104$ [$-0.0130,-0.0079$] \textsc{contr.} & $-0.0068$ [$-0.0094,-0.0041$] \textsc{contr.} & $-0.0022$ [$-0.0047,+0.0003$] \textsc{unres.} \\
E$^{\ddagger}$ & $-0.0136$ [$-0.0183,-0.0091$] & $-0.0094$ [$-0.0142,-0.0048$] & $-0.0037$ [$-0.0081,+0.0006$] \\
F$^{\ddagger}$ & $-0.0126$ [$-0.0162,-0.0090$] & $-0.0084$ [$-0.0121,-0.0048$] & $-0.0028$ [$-0.0064,+0.0008$] \\
Terra$^{\S}$ & $-0.0069$ [$-0.0106,-0.0032$] & $-0.0038$ [$-0.0078,+0.0001$] & $-0.00002$ [$-0.0042,+0.0041$] \\
\midrule
\multicolumn{4}{@{}l}{\emph{R2 $\Delta$ [95\% CI]}}\\
A & $+0.0575$ [$-0.0413,+0.1509$] & $+0.1662$ [$+0.0662,+0.2617$] & $+0.2837$ [$+0.1814,+0.3867$] \\
B & $+0.1473$ [$+0.0493,+0.2392$] & $+0.2644$ [$+0.1702,+0.3564$] & $+0.4222$ [$+0.3260,+0.5208$] \\
C & $-0.1199$ [$-0.1416,-0.0979$] & $-0.0849$ [$-0.1063,-0.0632$] & $-0.0432$ [$-0.0627,-0.0241$] \\
D$^{\dagger}$ & $-0.1028$ [$-0.1269,-0.0783$] & $-0.0698$ [$-0.0935,-0.0455$] & $-0.0242$ [$-0.0448,-0.0026$] \\
E$^{\ddagger}$ & $-0.1307$ [$-0.1725,-0.0906$] & $-0.0897$ [$-0.1295,-0.0511$] & $-0.0373$ [$-0.0718,-0.0026$] \\
F$^{\ddagger}$ & $-0.1233$ [$-0.1583,-0.0893$] & $-0.0836$ [$-0.1181,-0.0498$] & $-0.0253$ [$-0.0561,+0.0057$] \\
Terra$^{\S}$ & $-0.0681$ [$-0.1024,-0.0344$] & --- & --- \\
\bottomrule
\end{tabular}
\end{center}

\footnotesize $^{\dagger}$Descriptive only: cell D ran in the immediately
preceding serving window and completed before the A/B/C interleaved window
opened, so its outcome preceded the rule freeze.
$^{\ddagger}$Exploratory; no classification is conferred.
$^{\S}$Outside the factorial and its classification scheme; the
transport's frozen protocol reports R2 at the canonical cutoff only.
\normalsize

Two readings this table adds to the summary in
Section~\ref{sec:comparator}: cells E and F show the identical
descriptive pattern with no classification conferred, and F's is the only
R2 interval that spans zero at a stricter cutoff. As in the main text,
this is a descriptive re-analysis, not a preregistered interaction test.

\textbf{Support diagnostics.} Both endpoints are support-dependent
(Section~\ref{sec:measurement}), so the table below is required to read
their magnitudes. Support and residual statistics are reported over the
R2-scoreable population of each condition. For the cross-family transport
the two available populations differ and are therefore named separately:
mean named-candidate support is $8.3246$ over its eligible belief plies
($n=1{,}186$, the value cited in the main text) and $8.2691$ over its
R2-scoreable subset ($n=1{,}167$); the true state falls in the unnamed
residual on $1.77\%$ of eligible belief plies and $1.80\%$ of R2-scoreable
plies, and mean residual mass is $0.00447$. Neither pair may be reported
without its population.

\begin{center}
\small
\begin{tabular}{@{}lrrrr@{}}
\toprule
Condition & R2-scoreable plies & mean support $k$ & median $k$ & true state in OTHER \\
\midrule
Historical (original, 20 games) & 202 & 3.015 & --- & 42.08\% \\
Historical (replication, 60 games) & 514 & 2.938 & --- & 55.06\% \\
A & 1{,}439 & 4.499 & 4 & 23.00\% \\
B & 1{,}458 & 4.485 & 4 & 30.80\% \\
C & 2{,}404 & 8.395 & 9 & 1.46\% \\
D & 2{,}307 & 8.681 & --- & 1.08\% \\
E & 1{,}133 & 8.168 & 8 & 1.15\% \\
F & 1{,}119 & 8.458 & 9 & 1.25\% \\
\bottomrule
\end{tabular}
\end{center}

\footnotesize In the median-$k$ column, --- denotes a statistic not
reported in the frozen artifact; no post-hoc value was recomputed. Every
other cell in this table is a reported value, and no zero is represented
by a dash. \normalsize

\textbf{Capture-time diagnostics.} A separate population again: one event
per named candidate at capture-time plies, scored as the ratio of model to
matched-uniform Brier loss, with high-confidence commitments counted on
the captured square.

\begin{center}
\small
\begin{tabular}{@{}lrrrr@{}}
\toprule
Condition & captures & Brier ratio & commitments at $p\geq0.5$ & correct \\
\midrule
Historical (original, 20 games) & 99 & 7.127 & 22 & 0 \\
Historical (replication, 60 games) & 295 & 7.094 & 40 & 1 \\
A & 811 & 4.604 & 120 & 8 \\
B & 841 & 4.157 & 115 & 12 \\
C & 1{,}940 & 0.992 & 106 & 65 \\
D & 1{,}849 & 0.973 & 96 & 57 \\
E$^{\ddagger}$ & 930 & 0.874 & 45 & 34 \\
F$^{\ddagger}$ & 943 & 0.889 & 39 & 25 \\
\bottomrule
\end{tabular}
\end{center}

\footnotesize The cross-family transport is deliberately absent from this
table. Its frozen protocol defines its capture diagnostics over captures
of the \emph{true Regent square} rather than over any capture, giving a
proper-score ratio of $1.635$ on $67$ such captures; that quantity and the
column above are different populations and must not be placed on one
axis. \normalsize

\subsection{Objective state-richness stratification}
\label{app:kstrat}

Detail supporting Section~\ref{sec:comparator}. This analysis is
exploratory relative to the frozen factorial design and was specified in
full before any stratified value was computed.

\textbf{Definition.} $K$ is the number of live opponent pieces at the
scored ply, read from public game state immediately before that ply's
move. It is not a new quantity: $K$ is already the denominator of the
frozen matched-uniform comparator, so every eligible R1 event carries it.
Under the environment's rules a Crown Shift may move royal status to any
live own piece with no piece-type restriction, so the set of squares that
could hold the opponent's Regent is exactly the set of live opponent
pieces. No hidden or model-reported information enters $K$.

\textbf{Focal stratum and support rule.} The focal stratum is $K \geq 8$
--- at least half of an opponent's starting pieces still live --- which
covers over 80\% of eligible events in each core cell. Support is judged
by the study's existing convention: fewer than 100 contributing game
clusters is \textsc{instrument-limited} and reported descriptively rather
than inferentially.

\begin{center}
\small
\begin{tabular}{@{}llrrl@{}}
\toprule
Cell & $K\geq8$ $\Delta_{R1}$ [95\% CI] & events & clusters & support \\
\midrule
A & $+0.0147$ [$+0.0002,+0.0279$] & 5{,}351 & 145 & clears floor \\
B & $+0.0200$ [$+0.0064,+0.0331$] & 5{,}801 & 147 & clears floor \\
C & $-0.0130$ [$-0.0157,-0.0103$] & 17{,}473 & 150 & clears floor \\
D & $-0.0110$ [$-0.0139,-0.0082$] & 17{,}942 & 150 & clears floor \\
E & $-0.0131$ (no interval reported) & 8{,}474 & 75 & \textsc{instrument-limited} \\
F & $-0.0127$ (no interval reported) & 8{,}901 & 75 & \textsc{instrument-limited} \\
\bottomrule
\end{tabular}
\end{center}

Cells E and F fall below the 100-cluster floor and are descriptive only;
no interval is presented for them and they form no part of the
inferential high-$K$ claim.

\textbf{Cross-identifier contrasts (earlier minus later tested
identifier), under R.} At $K\geq8$: $+0.0277$, 95\% CI
$[+0.0128,+0.0414]$ at 4K (A versus C) and $+0.0310$,
$[+0.0172,+0.0446]$ at 16K (B versus D). The corresponding all-$K$
contrasts are $+0.0373$, $[+0.0233,+0.0498]$ and $+0.0367$,
$[+0.0236,+0.0492]$. These contrasts are themselves new relative to the
frozen design, which specified no cross-cell test, and are exploratory on
the same terms as the stratification. The B-versus-D pairing additionally
crosses serving windows.

\textbf{Attenuation, and what it is not.} The high-$K$ contrasts are
modestly smaller than the all-$K$ contrasts --- approximately 26\% at 4K
and 16\% at 16K. Two qualifications apply and neither is optional. The
intervals overlap substantially, and \emph{no predeclared test establishes
the attenuation itself}; it is a difference between two exploratory
estimates, not a tested effect. The direction is the one the existing
depth signature already predicts, which is a consistency observation
rather than independent evidence.

\textbf{Classification.} K1: the reversal persists clearly at $K\geq8$
with support above the floor and intervals excluding zero, so late-game
candidate-set collapse cannot by itself explain it. The claim applies
inferentially to the Gemini 3.1 and 3.7 cells only, and does not establish
that state richness has no effect.

\textbf{Descriptive pattern, below the support floor.} Across the low
($K\leq4$), mid ($5\leq K\leq7$) and high strata, the Gemini 3.7 cells are
comparatively flat while more of the $K$-dependence appears in the
Gemini 3.1 cells, whose low and mid strata are their worst. Those strata
fall below the support floor, so this is context rather than a claim ---
and it runs opposite to the direction a simple depletion account requires.

\textbf{Independent verification.} $K$ was verified by a second derivation
rather than restated: the canonical path replays recorded moves and Crown
Shifts on the engine and counts live pieces, while the independent path
counts pieces in each ply record's own stored board state. Every cell's
full $K$ distribution matched exactly, as did every event and cluster
count, across 148 compared quantities, with no discrepancies beyond
floating-point summation order.

\subsection{Well-formedness and format integrity}
\label{app:wellformed}

\textbf{Definitions.} A belief response is \emph{well-formed} if it parses
to a valid candidate distribution obeying the frozen mass-validity
tolerance, which admits a total stated mass within $0.02$ of $1$;
\emph{incoherent} if it is parseable JSON that violates a candidate-set or
mass-tolerance constraint; and \emph{malformed} if no valid JSON is
recoverable at all. Truncation --- a call reaching the generated-token
ceiling --- is recorded on a separate axis, because a truncated call can
still parse as well-formed if a complete object was emitted before the
cutoff. Responses that are not well-formed are excluded from scoring
rather than repaired.

\textbf{Denominator.} The population is the condition's logged decision
plies, as defined by the frozen scorer's own integrity report: one record
per model decision ply, partitioned by that record's elicitation category.
This is the definitional denominator and is the one used below. A
numerically identical count is obtainable from the opportunity module,
which replays the recorded move list and emits one entry per
model-mover ply; the two agree in every condition because the harness
emits exactly one record per model decision ply, but only the former
defines the quantity.

\begin{center}
\small
\begin{tabular}{@{}lrrrrrrr@{}}
\toprule
Cell & decision plies & well-formed & incoherent & malformed & WF rate & belief trunc. & move trunc. \\
\midrule
A & 2{,}288 & 2{,}057 & 231 & 0 & 89.90\% & 0/2{,}288 & 0/2{,}288 \\
B & 2{,}335 & 2{,}080 & 255 & 0 & 89.08\% & 0/2{,}335 & 0/2{,}335 \\
C & 3{,}177 & 3{,}129 & 48 & 0 & 98.49\% & 0/3{,}177 & 2/3{,}177 \\
D & 3{,}096 & 3{,}033 & 62 & 1 & 97.97\% & 0/3{,}096 & 0/3{,}096 \\
E & 1{,}580 & 1{,}536 & 15 & 29 & 97.22\% & 384/1{,}580 (24.3\%) & --- \\
F & 1{,}572 & 1{,}549 & 20 & 3 & 98.54\% & 0/1{,}572 & --- \\
\bottomrule
\end{tabular}
\end{center}

Gemini 3.7's well-formed rate is materially higher than Gemini 3.1's ---
roughly nine points, C/D against A/B --- after normalising by decision-ply
count, so it reflects a lower failure rate and not merely a larger ply
population. Cell D's 63 non-well-formed beliefs were verified
individually: 62 are incoherent, a candidate-set or mass-tolerance
violation on an otherwise-parseable payload, and exactly one is malformed,
with no valid JSON recoverable despite a normal, non-truncated completion.
Stated with that exception: across all four core cells, every belief
failure but that single case was a semantic failure on an
otherwise-parseable payload, and no belief response was truncated in cells
A--D. Cell E is the one condition anywhere in this study where the
generated-token ceiling materially bound generation, and its well-formed
rate nonetheless remains high for the reason given above. Move calls were
100\% well-formed in every condition.

\subsection{Same-window measurement-configuration control}
\label{app:hrcontrol}

\textbf{Purpose.} The historical runs and their reconstruction differ in
serving period and in measurement configuration at once, so neither source
is identified by their difference alone. This control holds the model
identifier, ceiling, opponent, horizon, seed schedule and scorer fixed and
varies the configuration bundle, within one day.

\textbf{Held constant.} The same public model identifier; the same serving
window; a 4{,}096-token generated-output ceiling; the same fixed Tier-1
opponent; a 150-ply game horizon; one shared environment-seed schedule;
balanced 50/50 colour assignment; and the common R1 scorer. One hundred
games per arm.

\textbf{Configurations.} H is the historical harness and R the rebuilt
native-API instrument used for cells A--F. They differ as a six-component
bundle, and the control does not resolve them individually.

\begin{center}
\small
\begin{tabular}{@{}lll@{}}
\toprule
Component & H (historical) & R (rebuilt) \\
\midrule
Transport & OpenAI-compatible endpoint & native provider API \\
Elicitation mode & combined move+belief & split move/belief \\
Own-state presentation & disambiguated & default \\
Thinking setting & none & fixed medium \\
Sampling-parameter forwarding & temperature/top-$p$ forwarded & not forwarded \\
Rules-document revision & archived revision & current revision \\
\bottomrule
\end{tabular}
\end{center}

\textbf{The rules-document component.} This component could be mistaken
for a change in the game, so it is stated exactly. The two revisions
differ by a single hunk at the end of the file --- the document version
block --- and the difference contains no normative text. Gameplay rules
were therefore identical across the arms. The reconstruction reads the
archived revision, verified by recomputing the document digest at read
time, and a counterfactual substitution of the current revision was run to
confirm the check is capable of failing. The rules text is read in a mode
that normalises line endings, so platform line-ending policy cannot
perturb the digests.

\begin{center}
\small
\begin{tabular}{@{}lrrl@{}}
\toprule
Quantity & $\Delta_{R1}$ & 95\% CI & classification \\
\midrule
H-current (2{,}444 events, 94 scoreable games) & $+0.0530$ & $[+0.0329,+0.0714]$ & \textsc{replicates} \\
R-current (4{,}404 events, 97 scoreable games) & $+0.0101$ & $[-0.0036,+0.0234]$ & \textsc{unresolved} \\
H$-$R configuration contrast (primary) & $+0.0429$ & $[+0.0182,+0.0667]$ & positive \\
Serving-period contrast (secondary) & $-0.0166$ & $[-0.0483,+0.0157]$ & inconclusive \\
Historical 60-game reference arm & $+0.0696$ & $[+0.0428,+0.0937]$ & \textsc{replicates} \\
\bottomrule
\end{tabular}
\end{center}

Both arms ran 100 games. The serving-period contrast is H-current minus
the historical 60-game arm. The H$-$R configuration contrast is primary:
within the same-day back-to-back comparison it measures the H-minus-R
endpoint difference,
while residual intra-window serving variation cannot be excluded. The
serving-period contrast compares one configuration across serving windows
and is secondary by design, because the historical arm is fixed at 60
games: its interval half-width cannot fall below $0.0254$ however much new
data is collected, and the realised half-width is $0.0320$. Its
inconclusiveness is therefore a statement about precision, not evidence of
equivalence. The 20-game historical arm ($+0.0593$,
$[+0.0292,+0.0931]$) is reported only as descriptive sensitivity and is
never pooled with the 60-game arm.

\textbf{Integrity.} Both arms completed 100 of 100 games on the frozen
seed schedule, with no errored games, no game reaching the 150-ply
horizon, and no truncated generations; every configuration field was
single-valued within each arm. An independent reimplementation sharing no
scoring code reconciled every compared quantity, with differences confined
to floating-point summation order.

\textbf{Format-level behaviour.} Well-formed elicitation rates were not
identical across serving windows even under H: 82.3\% in the archived
historical arm against 72.1\% (1{,}212 of 1{,}682) for H-current, a
difference of $-10.3$ percentage points, inside the prospectively frozen
$\pm15$-point diagnostic band. This records that format-level behaviour
differed across windows under one configuration. It is not evidence of a
mechanism, and not evidence of a backend change.

\subsection{Provenance milestones}
\label{app:provenance}

Because this paper argues that a nominal model identifier is insufficient
provenance, the execution record is made auditable here.

\begin{center}
\small
\begin{tabular}{@{}llr@{}}
\toprule
Condition & Public API model identifier & Games \\
\midrule
Historical (original) & \texttt{gemini-3.1-flash-lite} & 20 \\
Historical (replication) & \texttt{gemini-3.1-flash-lite} & 60 \\
D & \texttt{gemini-3.7-flash} & 150 \\
A & \texttt{gemini-3.1-flash-lite} & 150 \\
B & \texttt{gemini-3.1-flash-lite} & 150 \\
C & \texttt{gemini-3.7-flash} & 150 \\
E & \texttt{gemini-3.8-flash} & 75 \\
F & \texttt{gemini-3.8-flash} & 75 \\
Terra & \texttt{gpt-5.6-terra} (OpenAI Chat Completions) & 75 \\
\bottomrule
\end{tabular}
\end{center}

\textbf{Execution and freeze order.} Historical runs $\rightarrow$ cell D,
which motivated the grid $\rightarrow$ cells A/B/C, frozen and sealed
before any of their scientific outcomes were known and executed on
interleaved schedules within one serving window $\rightarrow$ the
Gemini 3.8 extension design freeze $\rightarrow$ cell E $\rightarrow$ a
sample-size and status amendment $\rightarrow$ cell F $\rightarrow$ the
cross-family transport, prospectively frozen and executed after the Gemini
conditions concluded. Cell D ran in the immediately preceding serving
window and completed before the A/B/C interleaved window opened.

\textbf{The extension amendment.} The extension was initially frozen at
150 games per cell, before any call to the new identifier. After cell E's
first 75-game block completed, only prespecified operational information
--- runtime, cost, transport behaviour and output-ceiling pressure --- was
inspected; no scientific endpoint, belief content or correctness signal
from either cell was inspected or scored. On that operational basis alone,
and before cell F began, the design was amended to 75 games per cell and
downgraded to exploratory. Cell E therefore ran 75 games under the
original confirmatory plan and cell F entirely under the amended
exploratory one.

\textbf{Transport configuration.} The cross-family transport was accessed
through the OpenAI Chat Completions API at reasoning effort \emph{high}
and the provider-default temperature of 1.0, with a 16{,}384-token
generated-output ceiling, the same fixed Tier-1 opponent and split
move/belief elicitation architecture as the core grid, and the same
canonical 75 environment seeds already used for cell F. Its sample size
was fixed prospectively rather than by an outcome-dependent stopping rule.

\textbf{The identifier interval.} The historical and reconstructed
Gemini 3.1 conditions share the identical public model identifier across
an 18--21 day interval, the historical runs preceding the reconstructed
ones. This is the paper's central provenance point, stated auditably.

\textbf{Metadata-resolution audit asymmetry.} Only the Gemini 3.8
identifier received an explicit provider model-metadata resolution audit;
the provider-returned resolved version string was \texttt{3.0}, carrying
no preview or experimental marker. No equivalent resolved-version audit
exists for \texttt{gemini-3.1-flash-lite} or \texttt{gemini-3.7-flash} in
any execution window, including the historical runs and their
reconstruction. This is an asymmetry in how firmly the served endpoint
behind each identifier was verified. It is not evidence that any
identifier resolved to something unexpected: nothing here indicates the
unaudited identifiers were wrong, only that our records cannot demonstrate
what they resolved to. It is a real provenance gap, and precisely the
condition under which this paper's snapshot-immutability limitation
applies.

\textbf{Anonymity.} Repository, commit and account identifiers, and
internal execution timestamps, are omitted throughout. The sequencing
above is the scientifically load-bearing content and is stated exactly.

\subsection{Elicitation-format feasibility studies}
\label{app:elicitation}

Detail supporting Section~\ref{sec:elicitation}. Three protocols were
tested before the factorial grid, each prospectively frozen with its own
read order, classification rule and prohibited-quantity list, all against
the fixed Tier-1 opponent with the evaluation design otherwise unchanged.
Belief elicitation is stateless, so both formats in a given comparison
could be issued at the identical pre-move state within one game, making
each comparison paired.

\textbf{Blinding.} Each protocol's frozen ruling was forbidden to inspect
Regent truth, correctness, proper scores or acted-on probability. These
studies therefore say nothing about what scientific belief-quality result
an exhaustive protocol would have produced; they speak only to whether it
could be elicited at all, and how stably.

\textbf{Protocol B: self-enumerated full support.} The model reconstructs
the live candidate set and assigns a probability to every member, with no
residual category. Non-inferiority margin $M=-0.15$, fixed before
execution.

\begin{center}
\small
\begin{tabular}{@{}llrrrll@{}}
\toprule
Band & required? & checkpoints & games & paired $\bar{D}$ & 95\% CI & verdict \\
\midrule
ply 1--25 & yes & 334 & 24 & $-0.6587$ & $[-0.7157,-0.6032]$ & \textbf{NO-GO} \\
ply 26--50 & yes & 60 & 3 & $-0.2000$ & $[-0.4000,\ 0.0000]$ & \textsc{unresolved} \\
ply 51+ & diagnostic & 74 & 2 & $-0.0811$ & $[-0.3333,+0.0638]$ & \textsc{unresolved} \\
\bottomrule
\end{tabular}
\end{center}

The ply 1--25 band is a required, well-powered band with all 24 games
contributing, and its entire interval lies below the margin: \textbf{NO-GO}.
The two later bands are \textsc{unresolved} at small sample sizes, on 3 and
2 contributing games respectively, so the formal decision rests on the
prespecified 1--25 band alone and not on any pooled result.
Well-formedness in that band fell from 90.4\% (302 of 334) under top-$k$
plus residual to 24.6\% (82 of 334) under full support.

\emph{Failure taxonomy.} Of 369 non-well-formed full-support responses,
182 named a candidate that was not live and 172 omitted one that was,
while only 15 violated the mass tolerance --- and there were zero
malformed responses in either arm. So 354 of the 369 failures are failures
to name the exact live candidate set. What collapsed was the model's
ability to enumerate that set, not its ability to produce structured
output.

\textbf{Protocol C: roster-supplied positional elicitation.} The harness
supplies the exact live-candidate roster and the model returns a
probability vector aligned to it. Structural feasibility improved
substantially relative to self-enumeration. A validation permuting only
roster order, with board state, history and candidate set held fixed,
then found that on every one of the 3 genuinely informative states,
cross-order variation exceeded same-order replicate variation, by factors
of $2.00\times$ to $15.00\times$. The frozen closure language is retained
rather than upgraded: the positional-vector representation exhibited
credible presentation-order sensitivity beyond ordinary same-condition
stochastic variation on informative states, and is therefore not
sufficiently stable for downstream belief-quality analysis under this
configuration; the roster-supply insight is retained, the positional
response encoding is not.

\textbf{Protocol D: identity-anchored roster supply.} Each probability is
bound explicitly to a named candidate rather than a vector position,
removing transcription between position and identity as a failure mode.
Two distinct instrument properties were assessed.

\emph{D-F1, structural feasibility} against concurrent top-$k$: mean
$D_{F1} = +0.2235$, 95\% CI $[+0.0824,+0.3647]$, against the same
$M=-0.15$ margin --- entirely above it, a formal \textbf{GO}. One
supporting statistic is retained: Protocol-D well-formedness on the
canonical call was 71 of 85.

\emph{D-F2, presentation stability}: mean total-variation distance between
responses to states differing only in hypothesis presentation order,
$0.1809$, 95\% CI $[0.1247,0.2395]$, achieved half-width $0.0574$ against
a frozen precision target of $0.10$ --- the declared precision target was
met. Five qualifications travel with this estimate.
(i) D-F2 was an \emph{estimation study with no prespecified pass or fail
threshold}; its interval lying above zero is evidence about the size of a
presentation effect, not a verdict against a criterion. It is therefore
\emph{not} a failed test, and no threshold is supplied after the fact in
order to make it one.
(ii) It was computable on 49 of 85 formal states, 57.6\%, because it
required four independently well-formed responses per state; the
complementary 36 states were unscoreable.
(iii) The estimate is conditional on that all-four-well-formed subset and
should not be assumed to characterise the full formal-state population.
(iv) No direction of that selection effect is identified: $0.1809$ is
\emph{not} a lower bound or a floor, and no separate analysis establishes
that the unscoreable states would have shown greater instability.
(v) It does not reverse D-F1, which tests a different instrument property
--- the structural feasibility of one canonical call, versus the stability
of the full response under an irrelevant change of presentation. A single
call can be reliably well-formed while repeated exhaustive elicitation at
the same state remains brittle.
Signs across the 49 scoreable states were 35 positive, 9 zero and 5
negative.

\textbf{Scope rule.} Development stopped at Protocol D under the
prospectively frozen scope rule; no fourth protocol was designed. On
review, the residual presentation sensitivity was judged too large to
treat identity-anchored full support as sufficiently stable to carry
downstream belief-quality analysis under the tested configuration. The
protocols characterise one model and one configuration; a different model
might enumerate reliably.

\subsection{Within-choice belief-action alignment ancillary}
\label{app:withinchoice}

Before the factorial grid was designed, a descriptive analysis on the
historical corpus asked whether the captured square received more stated
probability than explicitly named, unchosen, capturable alternatives on
the same ply. Eligibility required both the selected target and at least
one capturable alternative to be explicitly named, which only 32 plies
across 23 games satisfied. Within that subset, the captured square
received on average $+0.2184$ more stated probability than the mean named
unchosen alternative.

Two restrictions bound the reading. The subset is unrepresentative of
capture behaviour by construction: 73.6\% of historical captures targeted
a square never named at all, and those plies cannot enter the comparison.
And requiring alternatives to be named fixes the choice set, which makes
the comparison well-defined but does not establish why the model chose as
it did --- a model may capture a piece for material reasons while also
having assigned it more Regent probability than the other capturable
alternative. The defensible reading is only that, among explicitly
represented capturable alternatives, the chosen target tended to receive
higher stated Regent probability than the unchosen ones. It is not a
claim about why the choice was made, and not a calibration claim. The
opportunity-conditioned analysis of
Appendix~\ref{app:opportunity} supersedes this measure conceptually, by
conditioning on a population where belief is unambiguously actionable and
the relevant action is unambiguously relevant; this measure is retained
because it is the historical one and its restriction is instructive.

\subsection{Public-information reference tracker}
\label{app:pubtracker}

Historical and auxiliary context. This material is not part of the
factorial result reported in Section~\ref{sec:persistence}, and nothing in
the main text depends on it.

\textbf{Mechanics.} A predecessor study built a hand-specified belief
tracker that replays a recorded game and maintains a distribution over
which opponent piece holds royal status, updated only by deductively valid
rules keyed to observable game events: a check response on the Original
King; elimination of a captured piece when the game did not end; castling;
and the pre-shift timing window within which royal status cannot yet have
moved. The tracker never receives the opponent's true Regent identity;
ground truth enters only as the label against which its predictions are
scored. One implementation detail is worth stating plainly: the capture
rule is gated on a stored field recording whether a capture ended the
game, so the code reads a ground-truth value --- but what that field
stands for, namely that the game ended, is publicly observable, and no
hidden-state information is gained from it. Where this tracker is
described elsewhere as certificate-based, that refers to the logical
validity of the individual rules, not to an exhaustive predictor.

\textbf{Results.} Scored against the matched-uniform comparator on the
historical capture-only population at board ply $\geq 8$, by paired
bootstrap on the identical events:

\begin{center}
\small
\begin{tabular}{@{}lrrl@{}}
\toprule
Batch & $n$ & $\Delta$ (tracker $-$ uniform) & 95\% CI \\
\midrule
Original & 77 & $-0.00061$ & $[-0.00118,-0.00012]$ \\
Replication & 220 & $-0.00023$ & $[-0.00079,+0.00040]$ \\
\bottomrule
\end{tabular}
\end{center}

The original batch shows a small advantage whose interval excludes zero
marginally; the replication, which is the better-powered of the two, shows
an interval that spans zero. Against an underlying Brier scale of roughly
$0.017$--$0.024$ both point estimates are tiny, on the order of one to
three percent in relative terms, and they are reported as such rather than
as a robust demonstration in either direction.

\textbf{A withdrawn interpretation.} An earlier interpretation treated
this tracker as an existence proof that the hidden state is predictable at
these moments. \emph{That interpretation was withdrawn}, and the reason is
specific: later inspection found that the per-ply certificate overlay was
not current at any of the scored capture queries, so the realised tracker
had considerably less power than the phrase ``exact tracker'' would
suggest, and its high-confidence predictions arose from structural
elimination alone. The strongest defensible reading is narrow: on this
opponent and this population, a public-information tracker of this design
obtains very little predictive advantage. It is used in
Section~\ref{sec:comparator} only as limited-headroom context, and not as
a stronger comparator, not as a tractability certificate, and not as
evidence of post-shift predictability.

\textbf{Engine-defect insulation.} A later engine audit found a stale
post-terminal Regent pointer after the winning move was pushed. This
result is unaffected, and the reason rather than the assertion is what
matters: the tracker records the true Regent label \emph{before} the
terminal move is pushed, and does not consume the affected field.

\textbf{Predictability and calibration are different questions.} Where
public evidence is weak, appropriate uncertainty is itself the calibrated
response, so low raw predictability does not make
confidence-versus-outcome scoring meaningless. The opponent's shift timing
is additionally only partly state-responsive by policy
(Section~\ref{sec:design}), which bounds what any public-information
predictor could achieve on this population.

\subsection{Numerical audit note}
\label{app:audit}

Every reported number in this manuscript was checked against its source
scoring artifact. Most are stored fields checked by direct read: R1 and R2
deltas and intervals at all three depth cutoffs for A--D, E/F and the
transport; the candidate-support and capture diagnostics; the
opportunity-analysis tables; and the historical comparison table. A
systematic audit checked every A--F R1/R2 estimate and interval bound, at
all three reported cutoffs, against the frozen artifacts, and the
manuscript and its figure sources report the canonical artifact values
throughout.

One pair required reconstruction rather than a direct field read. Cell D's
absolute R2 Brier values --- model $0.7375$ and matched uniform $0.8403$,
quoted in Section~\ref{sec:comparator} --- are not themselves returned by
the frozen R2 scoring function, which reports only the difference. They
were recovered by re-running the frozen per-ply R2 formula over the D
corpus and separately averaging the model and uniform components before
differencing: recovered model $0.7375155537494574$ and uniform
$0.840299407246435$ on $n=2{,}307$ scoreable plies, matching the reported
figures and reproducing the already-frozen delta of $-0.102784$ exactly.

One presentational point is recorded for completeness. The historical
20-game arm's endpoint is $+0.0593$ at four decimal places, the direct
rounding of its stored value, and that form is used at every occurrence in
this paper.

\subsection{Opportunity-conditioned analysis: full cross-tabulations}
\label{app:opportunity}

\textbf{Definition.} An opportunity exists on a ply, for the model's own
turn, if and only if the opponent is in Regent Mode, the true Regent
square is known to the engine, and at least one legal move captures it.
The analysis is only possible while the opponent is in Regent Mode, so the
pre-shift window contributes zero opportunities by construction.
\emph{Converted} means the chosen move is exactly such a capture. This
analysis is ancillary and descriptive throughout: it introduces no
inferential test and does not modify R1 or R2.

The historical within-choice analysis of
Appendix~\ref{app:withinchoice} asks a related question on a differently
constructed population: it is restricted to plies where the eventual
capture target was itself among the model's named candidates, whereas the
analysis here conditions on positions in which belief is unambiguously
actionable, whether or not the true state was named.

\textbf{Provenance differs by condition}, and the difference is material
to how each row reads. The historical corpus and cells A--D were scored
post-hoc, under a specification written after those results already
existed. The E/F specification was frozen before scientific unseal ---
outcome-blind at freeze, though not prospective in the strong sense the
core grid's endpoints were. The transport's opportunity analysis was
prospectively frozen before execution.

All values below are from the corrected implementation described under
\emph{En-passant semantic correction}.

\begin{center}
\resizebox{\linewidth}{!}{%
\scriptsize
\begin{tabular}{@{}lrrrrrrrr@{}}
\toprule
Condition & opp.\ plies & games w/ opp. & converted & usable belief & named & top-ranked & $p\geq0.5$ & conv.\ $\mid p\geq0.5$ \\
\midrule
Historical (both runs) & 62 & 20/80 & 7 (11.3\%) & 34 & 14 (41.2\%) & 6 (17.6\%) & 6 (17.6\%) & 1/6 \\
A & 145 & 39/150 & 25 (17.24\%) & 119 & 86 (72.3\%) & 17 (14.3\%) & 13 (10.9\%) & 8/13 \\
B & 152 & 50/150 & 26 (17.11\%) & 113 & 78 (69.0\%) & 25 (22.1\%) & 24 (21.2\%) & 12/24 \\
C & 335 & 148/150 & 148 (44.18\%) & 334 & 327 (97.9\%) & 102 (30.5\%) & 70 (21.0\%) & 65/70 \\
D & 347 & 150/150 & 150 (43.23\%) & 344 & 343 (99.7\%) & 108 (31.4\%) & 63 (18.3\%) & 57/63 \\
E & 179 & 74/75 & 74 (41.34\%) & 170 & 170 (100.0\%) & 59 (34.7\%) & 38 (22.4\%) & 34/38 \\
F & 196 & 75/75 & 75 (38.27\%) & 196 & 194 (99.0\%) & 55 (28.1\%) & 32 (16.3\%) & 25/32 \\
Terra & 182 & 68/75 & 67 (36.81\%) & 173 & 169 (97.7\%) & 28 (16.2\%) & 29 (16.8\%) & 23/29 \\
\bottomrule
\end{tabular}%
}
\end{center}

Named, top-ranked and $p\geq0.5$ rates are over the usable-belief
population; conversion is over opportunity plies; and
conversion-given-$p\geq0.5$ is over the smaller population that reached
that threshold. In the newer conditions the true Regent is represented on
97.9--100\% of usable-belief opportunity plies, top-ranked on
28.1--34.7\% and assigned $p\geq0.5$ on 16.3--22.4\%; once that threshold
is reached the available win is usually taken. The co-variation with
tested identifier does not identify a causal mechanism.

\textbf{En-passant semantic correction.} The frozen semantic definition of
an actionable opportunity has always been that there exists a legal move
capturing the true Regent. A post-unseal audit found that the canonical
implementation approximated this as move-destination-square equality,
which is correct for every ordinary capture and capturing promotion but
misses en-passant captures, whose destination differs from the captured
piece's own square. Correcting the implementation changed the opportunity
denominator by exactly one ply each in cells C, D and F
($334\rightarrow335$, $346\rightarrow347$, $195\rightarrow196$) and the
conversion numerator by one ply in cell C only ($147\rightarrow148$, an
already-converted capture newly recognised as an opportunity rather than
newly converted); cells A, B, E and the transport were unaffected. The
three affected plies were individually confirmed to change no rate in
direction, since none was top-ranked or stated at $p\geq0.5$ under either
filter. Primary R1 and R2 endpoints were not recomputed and are
unaffected, because the opportunity module is not part of their scoring
path. Regression tests cover the en-passant case, an ordinary capture, a
capturing promotion, and castling, which can never register as a capture
under either filter. The original result artifacts were retained
unchanged, with the correction recorded separately, and every value in
this subsection and in Figure~\ref{fig:opportunity} is the corrected one.

\textbf{Post-hoc gameplay and material-depletion context.} Across the 825
games in the core grid, extension and transport combined, 822 ended by a
Regent capture on either side and 3 by repetition draw, with zero games
ending by checkmate after canonical classification. Terminal Regent
captures are late events: the median capture ordinal is the 13th--14th of
the winning side's captures, with a median of 3--4 opponent pieces
remaining immediately before it, the about-to-be-captured Regent included.
Winning trajectories nonetheless contain few genuine actionable
opportunities --- a median of 2--3 per winning game, with 1--2 missed
before the terminal capture. Read together, these are not the signature of
many repeated chances being squandered until the board clears: most
non-Regent captures occur in phases where the true Regent is not yet an
actionable target at all, and once an actionable opportunity exists it is
usually taken quickly. This is a post-hoc descriptive characterisation,
not a prespecified endpoint, and it does not establish robustness to
opponent strength.

\textbf{Opponent-regime engineering check.} After the primary experiments,
engine-only checks of stronger information-symmetric opponent
configurations were run, against frozen acceptance bands. In a
designated-side medium-versus-medium reference replay, 60\% of games
contained at least one actionable true-Regent opportunity; this fell to
8\% against a Tier-1 Hard configuration and to 3--4\% against the tested
Tier-2 search configurations, despite at least 99.5\% of games reaching
ply 14 and Crown Shifts occurring in all games. Trajectory length and
hidden-state activity therefore do not guarantee adequate opportunity
density. These were post-hoc engineering checks, not study endpoints, and
no reported result depends on them. No tested stronger configuration
cleared both the strategic-challenge and opportunity-yield criteria; the
intermediate design space remains largely unexplored, and no monotonic
strength-to-opportunity relationship is claimed from three tested points.

\subsection{Independent reimplementation}
\label{app:independent}

Every reported quantity in this study was checked by an independent
reimplementation sharing no code with the canonical scoring pipeline,
built from the frozen endpoint definitions and specification documents
alone.

\begin{center}
\small
\begin{tabular}{@{}lrrrl@{}}
\toprule
Scope & compared & exact & within $10^{-6}$ & discrepant \\
\midrule
Prospective core cells (A--C) & 162 & 135 & 27 & 0 \\
Extension (E, F) & 104 & 86 & 18 & 0 \\
Opportunity-conditioned analysis & 189 & 189 & 0 & 0 \\
Cross-family transport & 34 & 34 & 0 & 0 \\
State-richness stratification ($K$) & 148 & 148 & 0 & 0 \\
\bottomrule
\end{tabular}
\end{center}

For the prospective core cells every confidence interval reproduced under
the same clustered bootstrap and seed. The 162 quantities compared there
are divided equally across cells A, B and C; the descriptively reported D
cell was reproduced separately under its own audit and is not among them.
Of those 162, 135 agreed bit-for-bit and 27 agreed to within the $10^{-6}$
tolerance without being bit-identical. Those 27 are the R1 delta and both
interval bounds at each of the three reported depth cutoffs in each of the
three cells; they differ by at most $1.1\times10^{-14}$, which is
floating-point summation-order noise, and every R2 quantity and every
classification field matched exactly. No reported figure at its printed
precision, no interval interpretation, no endpoint classification and no
conclusion changes. For the extension, an implementation that
reconstructed all board mathematics from primitives differed only on R1
deltas and interval bounds, by $10^{-15}$ to $5\times10^{-15}$, which is
floating-point summation-order noise; R2 matched exactly throughout. The
opportunity reimplementation reproduced all compared quantities exactly
across all seven conditions, including the full high-confidence case
listings. For the transport, a second implementation was built by a
separate development pass from the frozen protocol prose alone, sharing no
scoring code and independently replaying game legality rather than
importing the canonical opportunity module; two implementation-side bugs
were found during reconciliation, both in diagnostic computation and never
in the R1/R2 endpoints, which matched from their first comparison, and
both were root-caused and fixed before the result was finalised.
Independent replay verified board-state alignment on sampled plies with
zero mismatches throughout.

\end{document}